%% file: main.tex
\documentclass[10pt,twocolumn,letterpaper]{article}

\usepackage[pagenumbers]{iccv} 

\definecolor{iccvblue}{rgb}{0.21,0.49,0.74}

\usepackage{algorithm}
\usepackage{etoolbox}
\usepackage{multirow}
\usepackage [accsupp]{axessibility}

\makeatletter
\AfterEndEnvironment{algorithm}{\let\@algcomment\relax}
\AtEndEnvironment{algorithm}{\kern2pt\hrule\relax\vskip3pt\@algcomment}
\let\@algcomment\relax
\newcommand\algcomment[1]{\def\@algcomment{\footnotesize#1}}
\renewcommand\fs@ruled{\def\@fs@cfont{\bfseries}\let\@fs@capt\floatc@ruled
  \def\@fs@pre{\hrule height.8pt depth0pt \kern2pt}%
  \def\@fs@post{}%
  \def\@fs@mid{\kern2pt\hrule\kern2pt}%
  \let\@fs@iftopcapt\iftrue}
\makeatother
\usepackage{listings}
\usepackage{overpic}
\usepackage{float}
\usepackage{graphicx}
\usepackage{rotating}
\usepackage{diagbox}
\usepackage{ragged2e}
\usepackage{microtype}
\usepackage{lipsum} 
\usepackage{diagbox}
\usepackage{subcaption}
\usepackage{pifont}
\usepackage[pagebackref,breaklinks,colorlinks,allcolors=iccvblue,urlcolor=purple]{hyperref}

\usepackage{colortbl} 
\definecolor{lightblue}{rgb}{0.97, 0.93, 0.96} 

\def\paperID{0} 
\def\confName{ICCV}
\def\confYear{2025}

\title{Reverse Convolution and Its Applications to Image Restoration}
\author{
  Xuhong Huang$^{1,}$\thanks{Equal contribution.} \quad
  Shiqi Liu$^{1,}$\footnotemark[1] \quad Kai Zhang$^{1,}$\thanks{Corresponding author (email: kaizhang@nju.edu.cn).} \quad Ying Tai$^1$ \quad Jian Yang$^1$ \quad Hui Zeng$^3$ \quad Lei Zhang$^{2,3}$\\
$^1$Nanjing University \quad  $^2$The Hong Kong Polytechnic University \quad $^3$OPPO Research Institute \\
\url{https://github.com/cszn/ConverseNet}
}

\begin{document}
\maketitle
\input{sec/0_abstract}    
\input{sec/1_intro}
\input{sec/2_related_work}
\input{sec/3_method}

\input{sec/4_implementation}

\input{sec/5_experiment}

\input{sec/6_conclusion}
{
    \small
    \bibliographystyle{ieeenat_fullname}
    \bibliography{main}
}

\end{document}

%% file: sec/0_abstract.tex
\begin{abstract}
Convolution and transposed convolution are fundamental operators widely used in neural networks. However, transposed convolution (a.k.a. deconvolution) does not serve as a true inverse of convolution due to inherent differences in their mathematical formulations. To date, no reverse convolution operator has been established as a standard component in neural architectures. In this paper, we propose a novel depthwise reverse convolution operator as an initial attempt to effectively reverse depthwise convolution by formulating and solving a regularized least-squares optimization problem. We thoroughly investigate its kernel initialization, padding strategies, and other critical aspects to ensure its effective implementation. Building upon this operator, 
we further construct a reverse convolution block by combining it with layer normalization, 1$\times$1 convolution, and GELU activation, forming a Transformer-like structure. The proposed operator and block can directly replace conventional convolution and transposed convolution layers in existing architectures, leading to the development of ConverseNet. Corresponding to typical image restoration models such as DnCNN, SRResNet and USRNet, we train three variants of ConverseNet for Gaussian denoising, super-resolution and deblurring, respectively. Extensive experiments demonstrate the effectiveness of the proposed reverse convolution operator as a basic building module. We hope this work could pave the way for developing new operators in deep model design and applications.
\end{abstract}

%% file: sec/1_intro.tex
\section{Introduction}
\label{sec:intro}

Convolution and transposed convolution (the latter sometimes referred to as deconvolution) are fundamental operations commonly used in deep neural networks. Convolution is used for feature extraction and can achieve downsampling to reduce spatial dimensions~\cite{lecun1998gradient}. In contrast, transposed convolution is widely used for upsampling the spatial dimensions of its inputs~\cite{chen2014semantic,noh2015learning,ronneberger2015unet,radford2015unsupervised}. 
This functional relationship has led some studies to regard transposed convolution as a form of reverse convolution. However, from a mathematical perspective, transposed convolution is not the true inverse of convolution; rather, it can be described as performing upsampling by inserting zeros between input elements, followed by a standard convolution.

\begin{figure}[tbp]
\centering
\subfloat[]
{\vspace{-0.1mm} \includegraphics[width=0.145\textwidth]{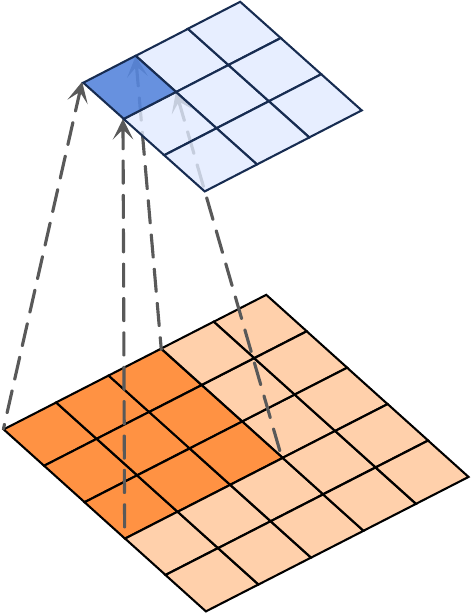}}
\hspace{-0.7mm}
\subfloat[]
{\vspace{1.3mm}
\includegraphics[width=0.145\textwidth]{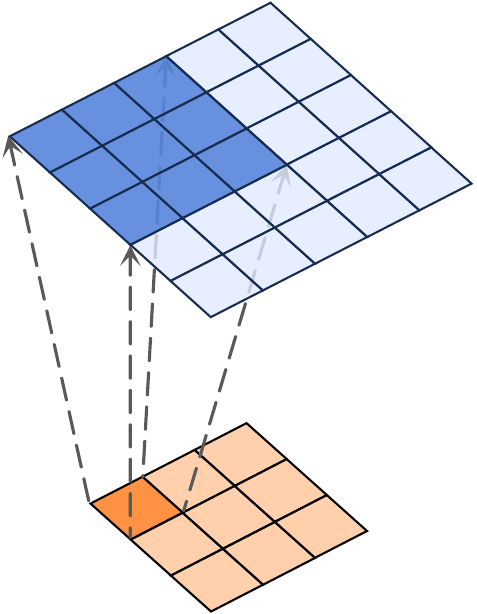}}
\hspace{1mm}
\subfloat[]
{\vspace{1.32mm}
\includegraphics[width=0.145\textwidth]{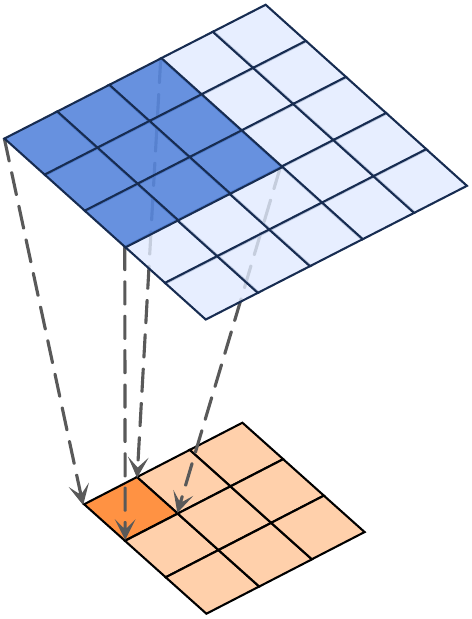}}
\caption{Illustration of the structural differences among (a) standard convolution, (b) transposed convolution, and (c) reverse convolution. Each subfigure shows a single input feature map (in orange) and its corresponding output feature map (in blue).} 
\label{fig:operators}
\end{figure}

One notable approach related to reverse convolution is deconvolution for image deblurring~\cite{schuler2013machine, son2021single,xu2014deep,kundur1996blind}, which aims to recover a sharp image from a blurred one by approximating the inverse of the blurring process. However, deconvolution methods typically involve iterative optimization~\cite{zhang2025blind,zhang2017learning,kundur1996blind,lee2021iterative} and are generally specialized for deblurring, limiting their flexibility as general neural network modules.
Another closely related work is invertible convolution~\cite{dinh2014nice, dinh2016density,kingma2018glow}, which ensures exact invertibility by imposing specific constraints, such as identical input-output shapes and a tractable Jacobian. These constraints restrict its applicability in deep networks. Notably, invertible convolution is essentially a constrained form of standard convolution, rather than a fundamentally new operator. Therefore, developing a true reverse convolution operator remains an open and valuable research direction.

In this paper, we make the first attempt to propose a reverse convolution operator, explicitly designed as the inverse of depthwise convolution. The main idea is to formulate the reverse convolution problem as the minimization of a quadratic regularized function, which enjoys a closed-form analytical solution.
Unlike deconvolution methods for image deblurring, which typically require iterative optimization and mainly operate on single-channel grayscale or three-channel RGB images with limited applications, our operator is computed in a single step and supports arbitrary channel dimensions, enabling broad adaptability across network architectures and tasks.
Unlike invertible convolution, which is primarily used in invertible neural networks or normalizing flows, our proposed operator offers broader applicability across various network structures.
Furthermore, similar to transposed convolution, our operator can perform spatial upsampling on feature maps. As illustrated in Fig.~\ref{fig:operators}, while standard and transposed convolution apply the kernel to the input, \textit{reverse convolution applies the kernel to the output to reconstruct the input}.

The proposed reverse convolution operator includes learnable kernel parameters and adjustable configurations such as kernel size and scaling factor.
To ensure effective implementation, we examine essential aspects such as kernel initialization and padding strategies. Building on this, we construct a reverse convolution block by combining it with layer normalization~\cite{ba2016layer}, $1\times1$ convolution, and GELU activation~\cite{hendrycks2016gaussian}, in a structure similar to a Transformer block~\cite{vaswani2017attention}.
We use this block to build a network, which we refer to as ConverseNet. In ConverseNet, the reverse convolution block is used to replace corresponding components in existing architectures such as DnCNN~\cite{dncnn} and SRResNet~\cite{srresnet}. We evaluate its effectiveness on representative image restoration tasks, including Gaussian denoising and super-resolution. Furthermore, due to its functional similarity to deconvolution, we incorporate it into USRNet~\cite{zhang2020deep} for deblurring, where experimental results demonstrate its clear advantages.

The main contributions of this work are as follows:
\begin{itemize} 
    \item 
    We propose a novel reverse convolution operator specifically designed as the mathematical inverse of depthwise convolution. By formulating the problem as a quadratic regularized minimization, we obtain a closed-form solution that enables non-iterative computation.
    \item We integrate the reverse convolution operator into a reverse convolution block, which serves as the fundamental building unit of ConverseNet. This design enables the network to explicitly capture spatial dependencies by incorporating the reverse convolution operator.
    \item We demonstrate the effectiveness of our reverse convolution operator through extensive experiments on typical image restoration tasks, including Gaussian denoising, super-resolution and deblurring. 
    \item We demonstrate that the proposed operator enables non-blind deblurring by directly conditioning on the blur kernel at the feature level, in contrast to most unfolding-based methods that operate in the single-channel grayscale or three-channel RGB image domain.
\end{itemize}

%% file: sec/2_related_work.tex
\section{Related Work}
\label{sec/2_related works}

\subsection{Convolution}

Convolution is a basic operator in deep neural networks, primarily used for feature extraction and spatial downsampling. Its key parameters such as kernel size and stride control the receptive field and resolution reduction, respectively.
To meet diverse computational and architectural demands, several convolution variants have been proposed.
Group convolution reduces computational cost by splitting channels into groups~\cite{ResNeXt, wang2019fully, su2020dynamic, huang2018condensenet}, while depthwise convolution improves efficiency by applying spatial convolution independently on each channel~\cite{hua2018pointwise, zhang2020high, tan2019mixconv,mobilenets, xception}.
Dilated convolution expands the receptive field by inserting zeros (dilations) between kernel elements, allowing the network to capture broader context without increasing the number of parameters~\cite{dilatedconvolution, khalfaoui2021dilated, chen2024frequency, gao2023rethinking}. 
Invertible convolution has been introduced in normalizing flow models~\cite{dinh2014nice, dinh2016density, kingma2018glow, kumar2019videoflow, DBLP:conf/iclr/GrathwohlCBSD19, DBLP:conf/nips/ChenBDJ19, DBLP:conf/icml/BehrmannGCDJ19}, but its adoption in general deep architectures is limited due to rigid shape constraints and mathematical restrictions.
Despite the progress in convolutional design, constructing an effective reverse convolution operator remains an open challenge.

\subsection{Transposed Convolution}

Transposed convolution is widely used for upsampling in encoder-decoder architectures~\cite{noh2015learning, ronneberger2015unet} and super-resolution tasks.
It increases spatial resolution by adjusting stride, but often introduces checkerboard artifacts due to uneven kernel overlap~\cite{odena2016checkerboard}. A common remedy is to use nearest-neighbor upsampling followed by a standard convolution.
Despite its name, transposed convolution is not a true reverse of standard convolution. It can be interpreted as inserting zeros between feature map elements, followed by a standard convolution.
However, designing a mathematically rigorous reverse convolution operator has received limited attention, and popular deep learning frameworks such as PyTorch lack native support for such functionality.

\subsection{Deconvolution}

Deconvolution generally aims to invert convolution operations and recover sharp images from blurred observations. Unlike transposed convolution, deconvolution is conceptually closer to reverse convolution.
Existing deconvolution methods can be broadly categorized into one-step and multi-step solutions.
Wiener deconvolution~\cite{wiener1949extrapolation} represents a classical one-step approach that provides a closed-form solution based on inverse filtering and regularization, assuming a known blur kernel.
In contrast, multi-step methods include plug-and-play frameworks~\cite{kamilov2017plug,zhang2021plug}, which incorporate pretrained denoisers into iterative solvers, and deep unfolding methods~\cite{zhang2020deep,wu2022uretinex,mou2022deep}, which unroll optimization procedures into end-to-end trainable networks.

While effective, these methods typically assume a known blur kernel and operate within the single-channel grayscale or three-channel RGB image domain, which limits their applicability to high-dimensional feature representations in modern deep networks. Although the deep Wiener deconvolution~\cite{dong2020deep} can be applied to deep features, it is primarily tailored for image deblurring. This limitation motivates the development of a more general reverse convolution operator that can be efficiently integrated into deep learning architectures.

%% file: sec/3_method.tex
\section{Method}

\subsection{Objective}

We aim to develop a depthwise reverse convolution operator capable of recovering the original input feature map from the output of a depthwise convolution operation followed by downsampling. Specifically, consider a single feature map $\mathbf{X} \in \mathbb{R}^{H \times W}$, where $H$ and $W$ represent the height and width of the feature map, respectively. This feature map is convolved with a kernel $\mathbf{K} \in \mathbb{R}^{k_h \times k_w}$, where $k_h$ and $k_w$ are the height and width of the kernel $\mathbf{K}$. The convolution is applied channel-wise, denoted by $\otimes$, and is followed by downsampling with a stride $s$, denoted by $\downarrow_{s}$. The resulting feature map is $\mathbf{Y} \in \mathbb{R}^{H' \times W'}$, where $H'$ and $W'$ are determined by the convolution and downsampling operations. Mathematically, this process is expressed as: 
\begin{equation} \label{eq:1} 
\mathbf{Y} = (\mathbf{X} \otimes \mathbf{K})\downarrow_{s}. 
\end{equation}

Our objective is to reconstruct $\mathbf{X}$ given the observed  $\mathbf{Y}$, the convolution kernel $\mathbf{K}$, and the stride $s$, which can be expressed as:
\begin{equation} \label{eq:2} 
\mathbf{X} = \mathcal{F}(\mathbf{Y}, \mathbf{K}, s).
\end{equation}
This involves inverting the combined effect of convolution and downsampling, which is a typical ill-posed problem.

\subsection{Problem Formulation}

To solve Eq.~\eqref{eq:2}, we formulate the recovery of $\mathbf{X}$ as an optimization problem that seeks an approximate solution by minimizing the error between the observed output $\mathbf{Y}$ and the reconstructed output $(\mathbf{X}\otimes\mathbf{K})\downarrow_{s}$. Specifically, we define the optimization problem as: 
\begin{equation} \label{eq:3} 
\mathbf{X}^\ast = {\arg\min}_{\mathbf{X}} \left\| \mathbf{Y} - \left( \mathbf{X} \otimes \mathbf{K} \right) \downarrow_{s} \right\|_F^2, 
\end{equation} 
where $\left\| \cdot \right\|_{F}$ denotes the Frobenius norm and is used to measure the error between the actual output $\mathbf{Y}$ and the output reconstructed by $(\mathbf{X}\otimes\mathbf{K})\downarrow_{s}$.

However, directly solving Eq.~\eqref{eq:3} may lead to unstable solutions due to the ill-posed nature of the problem. To stabilize the solution, we introduce a quadratic regularization term, giving rise to
\begin{equation} \label{eq:regularized} 
\mathbf{X}^\ast = {\arg\min}_{\mathbf{X}} \left\| \mathbf{Y} - \left( \mathbf{X} \otimes \mathbf{K} \right) \downarrow_{s} \right\|_F^2 + \lambda \left\| \mathbf{X} - \mathbf{X}_0 \right\|_F^2, 
\end{equation}
where $\lambda > 0$ is a regularization parameter controlling the trade-off between reconstruction and regularization, and $\mathbf{X}_0$ is an initial estimate for $\mathbf{X}$ that guides the solution towards a more stable and plausible result.
More precisely, the first term in the objective function ensures that when convolution and downsampling are applied to our estimated $\mathbf{X}$, the result approximates the observed $\mathbf{Y}$. The second term acts as a regularizer to prevent overfitting and promote solutions close to $\mathbf{X}_0$.

Up to this point, we have reformulated the original problem as finding an estimated feature map \(\mathbf{X}^\ast\) that minimizes the error between the observed \(\mathbf{Y}\) and the reconstructed output from \(\mathbf{X}^\ast\) after convolution and downsampling. This minimization is balanced by a regularization term that encourages \(\mathbf{X}^\ast\) to remain close to a prior estimation \(\mathbf{X}_0\).
Two simple choices for $\mathbf{X}_0$ are given by: 
\begin{equation} \label{eq:5} 
\mathbf{X}_0 = \begin{cases} \mathbf{0} \\ 
\text{Interp}(\mathbf{Y}, s), \end{cases}
\end{equation} 
where $\text{Interp}(\mathbf{Y}, s)$ denotes an interpolation operation applied to $\mathbf{Y}$ with a scaling factor of $s$.

By solving the optimization problem above, we aim to develop a depthwise reverse convolution operator that effectively inverts depthwise convolution with downsampling. This operator can serve as a modular component in deep neural networks, enhancing their capacity to reconstruct complex spatial features and enabling more flexible architectures.

\subsection{Closed-form Solution}

To solve the optimization problem in Eq.~\eqref{eq:regularized}, we can get a closed-form solution for $\mathbf{X}$ under the assumption of circular boundary conditions for the convolution~\cite{zhao2016fast}:
\begin{equation}\label{eq:solution}
  \mathbf{X}^{*} = \mathbf{F}^{-1}\!\left(\frac{1}{\lambda}\Big(\mathbf{L} - \overline{\mathbf{F}_K} \odot_{s}\frac{(\mathbf{F}_K\mathbf{L})\Downarrow_{{s}} }{|\mathbf{F}_K|^2\Downarrow_{{s}} +\lambda}\Big)\right),
\end{equation}
where $\mathbf{L} = \overline{\mathbf{F}_K}\mathbf{F}_{Y\uparrow_{s}} + \lambda\mathbf{F}_{X_{0}}$. Here, $\mathbf{F}{(\cdot)}$ denotes the fast Fourier transform (FFT), and $\mathbf{F}^{-1}(\cdot)$ represents the inverse FFT. The term $\overline{\mathbf{F}_K}$ indicates the complex conjugate of $\mathbf{F}_K$ for kernel $\mathbf{K}$, while $|\mathbf{F}_K|^2 = \overline{\mathbf{F}_K} \odot \mathbf{F}_K$ denotes the element-wise squared magnitude of the Fourier transform of the kernel $\mathbf{K}$. $\mathbf{F}_{Y\uparrow_{s}}$ represents the FFT of the standard $s$-fold upsampling $\mathbf{Y}$. The operator $\odot_s$ signifies element-wise multiplication applied to $s \times s$ distinct blocks, and $\Downarrow_s$ represents the distinct block downsampling operator that averages over these $s \times s$ blocks. The notation $\uparrow_s$ refers to the standard $s$-fold upsampling operator, which increases the spatial dimensions by inserting zeros. 
Note that Eq.~\eqref{eq:solution} holds under the assumption that $\downarrow_s$ denotes the standard $s$-fold downsampler, which selects the upper-left pixel from each $s\times s$ patch. When $s=1$, Eq.~\eqref{eq:solution} simplifies to:
\begin{equation}
\mathbf{X}^*  = \mathbf{F}^{-1} \left( \frac{ \overline{\mathbf{F}_K}\mathbf{F}_{Y} + \lambda \mathbf{F}_{X_0}}{ |\mathbf{F}_K|^2 + \lambda }\right).
\end{equation}
By applying the above closed-form solutions, we can effectively invert the convolution and downsampling operations to obtain $\mathbf{X}^\ast$. 
Notably, the kernel $K$ can either be learned jointly with the network in an end-to-end fashion, or specified as a known prior to serve as a conditioning input. It is also worth noting that the regularization parameter $\lambda$ can be jointly optimized during training, as demonstrated in Section~\ref{converse2d}, rather than being manually fixed as in most conventional deconvolution algorithms.

Following the naming convention of Conv2D, we refer to our reverse convolution operator as Converse2D. The PyTorch-like implementation is provided in Algorithm~\ref{algo}.

\subsection{Reverse Convolution Block}

To evaluate the effectiveness of the proposed depthwise Converse2D operator, we construct a converse block, drawing inspiration from the design principles of Transformer blocks. 
Since Converse2D operates in a depthwise manner and lacks inherent cross-channel interactions, we integrate $1 \times 1$ convolutions to enable channel mixing. 
As illustrated in Fig.~\ref{reverse_block}, the block also includes layer normalization, a GELU activation function, and residual connections.
In this configuration, the Converse2D operator is solely responsible for modeling spatial dependencies, while the $1 \times 1$ convolutions facilitate inter-channel information exchange. Such separation of spatial and channel-wise processing makes the Converse block an effective and flexible module for building deep networks.

Fig.~\ref{reverse_block} also shows three input feature maps (\ie, $\mathbf{Y}$), corresponding kernels (\ie, $\mathbf{K}$), and outputs (\ie, $\mathbf{X}^*$) of a Converse2D operator within $s=1$ in green dashed box, and reconstructed results (\ie, $\mathbf{X}^* \otimes \mathbf{K}$) and error maps (\ie, $\mathbf{Y}- \mathbf{X}^* \otimes \mathbf{K}$) in the gray dashed box. We can see Converse2D provides an effective solution to Eq.~\eqref{eq:3}, yielding $\mathbf{X}^*$ with enhanced structural details. This behavior is consistent with that of typical deconvolution operations.

%% file: sec/4_implementation.tex
\section{Implementation of Converse2D Operator}
\label{converse2d}

{\justifying As Converse2D involves several critical factors that can influence effectiveness and stability, we carefully implemented it based on the denoising task. In the following, we discuss these key components in detail: the kernel $\mathbf{K}$, the padding strategy, the regularization parameter $\lambda$, and the initial estimation $\mathbf{X}_0$. 

\vspace{0.2cm}
\noindent \textbf{Kernel $\mathbf{K}$:}
The initialization of the kernel $\mathbf{K}$ plays a critical role, as it directly influences the stability and accuracy of the solution produced by the Converse2D operator. Inspired by classical deblurring methods, we apply a Softmax normalization to $\mathbf{K}$ during pre-processing to enforce non-negativity and sum-to-one constraints.
Specifically, after random initialization, the kernel weights are normalized using the Softmax function:
\begin{equation} 
\mathbf{K}_{\text{normalized}} = \text{Softmax}(\mathbf{K}).
\end{equation}

\begin{algorithm}[!h]\small
    \caption{Code of Converse2D in a PyTorch-like style.}
    \label{algo}
    \definecolor{codeblue}{rgb}{0.25,0.5,0.5}
    \lstset{
        backgroundcolor=\color{white},
        basicstyle=\fontsize{7.2pt}{7.2pt}\ttfamily\selectfont,
        columns=fullflexible,
        breaklines=true,
        captionpos=b,
        commentstyle=\fontsize{8pt}{8pt}\color{codeblue},
        keywordstyle=\fontsize{8.0pt}{8.0pt},
    }
\begin{lstlisting}[language=python]
# X: input feature map
# K: kernel
# S: stride or upscaling factor
# p2o: convert point-spread function (PSF) to optical transfer function (OTF)
# splits: divide a tensor of shape [..., H, W] into [..., H/S, W/S, S^2] distinct blocks

B, C, H, W = X.shape        # X: B, C, H, W
# lambda: regularization parameter
lambda = torch.sigmoid(bias-9.0) + 1e-5
# upsample X: B, C, HxS, WxS
Y_S = upsample(X, scale=S)
# interpolate X: B, C, HxS, WxS
X_0 = nn.functional.interpolate(X, scale_factor=S, mode='nearest')

# FFT kernel: B, C, HxS, WxS
FK = p2o(K, (H*S, W*S))
# complex conjugate
FK_conj = torch.conj(FK)
# element-wise squared magnitude
FK_2 = torch.pow(torch.abs(FK), 2)
FKY = FK_conj*torch.fft.fftn(Y_S, dim=(-2, -1))
# L: B, C, HxS, WxS
L = FKY + torch.fft.fftn(lambda*X_0, dim=(-2,-1))
FKL = FK.mul(L)
# split and calculate mean: B, C, H, W
FKL_S = torch.mean(splits(FKL, S), dim=-1, 
      keepdim=False)
FK2_S = torch.mean(splits(FK_2, S), dim=-1, 
      keepdim=False)
Fdiv = FKL_S.div(FK2_S + lambda)
# element-wise multiplication
Fmul = FK_conj*Fdiv.repeat(1, 1, S, S)
Fout = (L-Fmul)/lambda
# inverse FFT output: B, C, HxS, WxS
out = torch.real(torch.fft.ifftn(Fout, dim=(-2, -1)))

return out
    \end{lstlisting}
\end{algorithm}

\vspace{-0.1cm}
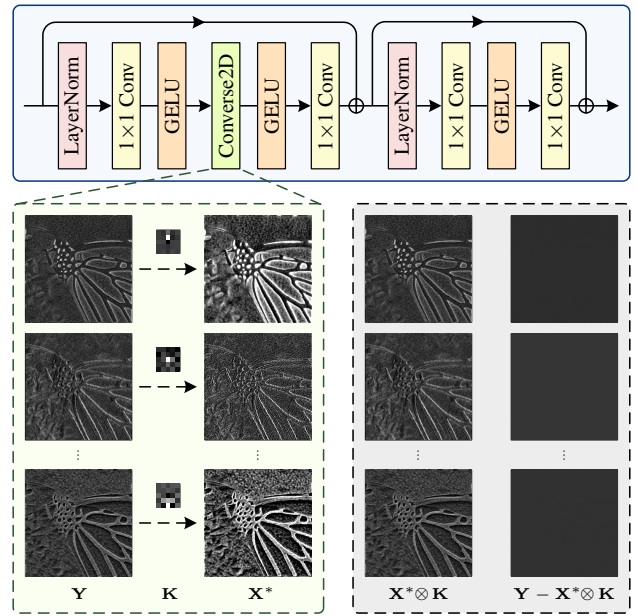
\begin{figure}[H]
\centering
\begin{overpic}[width=0.99\linewidth]{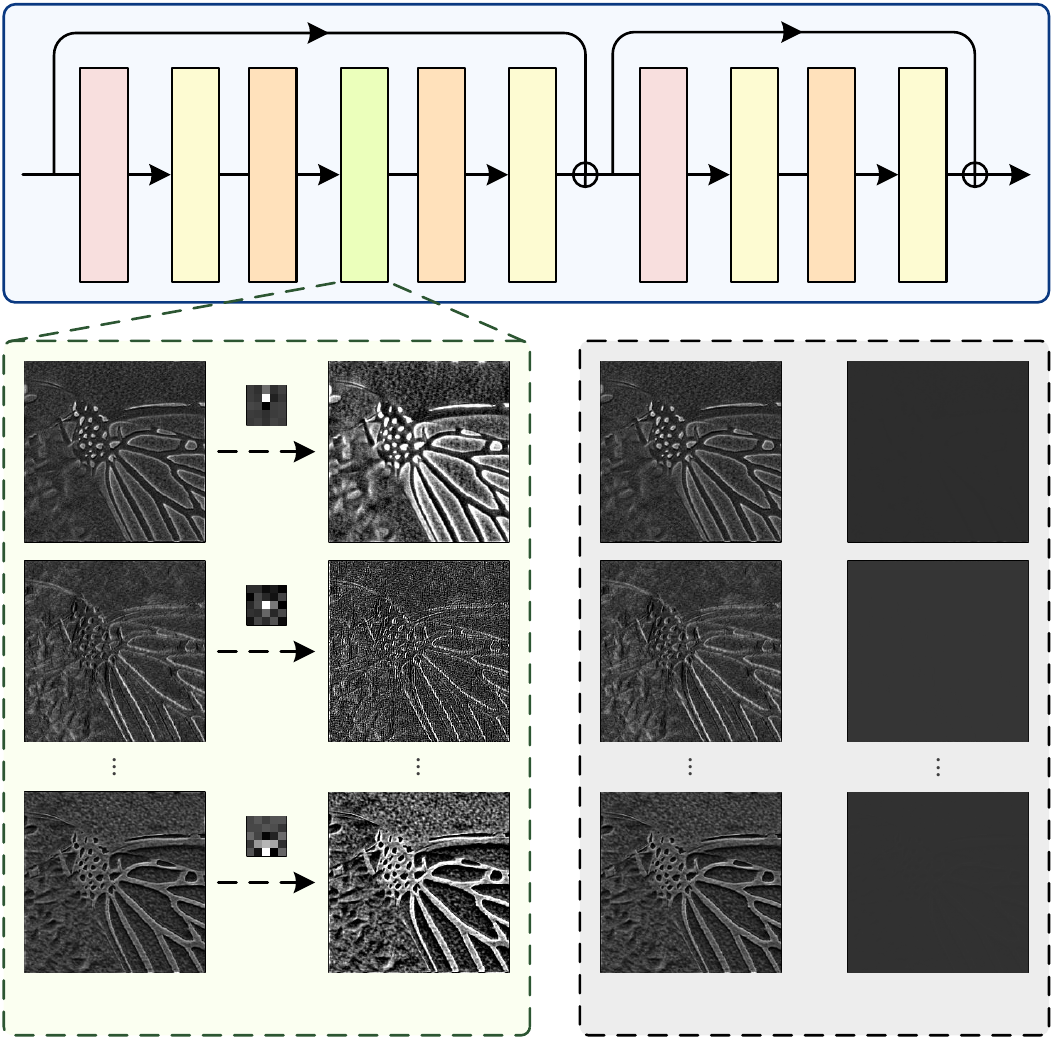}
    \put(8.6,74.29){\rotatebox{90}{\fontsize{8}{8}\selectfont\color{black}{LayerNorm}}}
    \put(17.4,74.94){\rotatebox{90}{\fontsize{8}{8}\selectfont\color{black}{1$\times$1 Conv}}}
    \put(24.79,77.7){\rotatebox{90}{\fontsize{8}{8}\selectfont\color{black}{GELU}}}
    \put(33.45,73.78){\rotatebox{90}{\fontsize{8}{8}\selectfont\color{black}{Converse2D}}}
    \put(40.84,77.7){\rotatebox{90}{\fontsize{8}{8}\selectfont\color{black}{GELU}}}
    \put(49.48,74.94){\rotatebox{90}{\fontsize{8}{8}\selectfont\color{black}{1$\times$1 Conv}}}
    \put(61.8,74.29){\rotatebox{90}{\fontsize{8}{8}\selectfont\color{black}{LayerNorm}}}
    \put(70.51,74.94){\rotatebox{90}{\fontsize{8}{8}\selectfont\color{black}{1$\times$1 Conv}}}
    \put(77.88,77.7){\rotatebox{90}{\fontsize{8}{8}\selectfont\color{black}{GELU}}}
    \put(86.51,74.94){\rotatebox{90}{\fontsize{8}{8}\selectfont\color{black}{1$\times$1 Conv}}}
    \put(9.68,2.65){\fontsize{6}{8}\selectfont\color{black}{$\mathbf{Y}$}}
    \put(24.23,2.65){\fontsize{6}{8}\selectfont\color{black}{$\mathbf{K}$}}
    \put(38.18,2.65){\fontsize{6}{8}\selectfont\color{black}{$\mathbf{X}^*$}}
    \put(60.83,2.65){\fontsize{6}{8}\selectfont\color{black}{$\mathbf{X}^* \quad \mathbf{K}$}}
    \put(80.99,2.65){\fontsize{6}{8}\selectfont\color{black}{$\mathbf{Y}- \mathbf{X}^* \quad \mathbf{K}$}}
    \put(91.65,2.85){\fontsize{6}{8}\selectfont\color{black}{$\otimes$}}
    \put(64.95,2.85){\fontsize{6}{8}\selectfont\color{black}{$\otimes$}}
\end{overpic}
    \caption{The architecture of the proposed reverse convolution block. The green dashed box shows three input feature maps (\ie, $\mathbf{Y}$), the corresponding kernels (\ie, $\mathbf{K}$), and the corresponding outputs (\ie, $\mathbf{X}^*$) of a Converse2D operator with $s=1$, while the gray dashed box shows the corresponding reconstructed results (\ie, $\mathbf{X}^* \otimes \mathbf{K}$) and error maps (\ie, $\mathbf{Y}- \mathbf{X}^* \otimes \mathbf{K}$).}
    \label{reverse_block}
\end{figure}

As shown in Table~\ref{table:kernel}, Softmax-based initialization outperforms both uniform and Gaussian initialization in the Gaussian denoising task.
Based on this observation, we adopt Softmax initialization in subsequent experiments and investigate the effect of kernel size. Results in Table~\ref{table:kernel} indicate that a $5 \times 5$ kernel offers a favorable trade-off between performance and model complexity.

\begin{table}[hpbt]
\caption{The average PSNR(dB) results of Converse-DnCNN with different kernel initialization methods and kernel sizes on Set12 and BSD68 datasets.}
\centering
\begin{minipage}{0.23\textwidth}
\raggedright
\footnotesize
\begin{tabular}{ccc}
\toprule
Initialization & \multicolumn{2}{c}{Datasets} \\
\cmidrule{2-3}
Methods & Set12 & BSD68 \\
\midrule 
Uniform & 30.31 & 28.92  \\
\midrule 
Gaussian & 30.60 & 29.30 \\
\midrule 
\rowcolor{lightblue}
Gauss.+Softmax & 30.70 & 29.36 \\
\bottomrule
\end{tabular}
\label{table_kernel_1}
\end{minipage}
\hspace{0.6cm}
\begin{minipage}{0.2\textwidth}
\raggedright
\footnotesize
\begin{tabular}{ccc}
\toprule
Kernel   & \multicolumn{2}{c}{Datasets} \\
\cmidrule{2-3} 
 Sizes&  Set12 & BSD68 \\
\midrule 
$3\times3$ &  30.58 &  29.29\\
\midrule 
$5\times5$ & 30.70 & 29.36 \\
\midrule 
\rowcolor{lightblue}
$7\times7$ & 30.71 & 29.36 \\
\bottomrule
\end{tabular}
\label{table_kernel_2}
\end{minipage}
\label{table:kernel}
\end{table}

\vspace{0.2cm}
\noindent \textbf{Padding Strategy:}
Padding affects how boundary information is treated during convolution and directly influences the spatial dimensions of the output. Therefore, it is important to investigate the impact of both padding mode and padding size in the proposed reverse convolution operator. The padding mode determines how values beyond the input boundaries are handled, with common choices including circular, reflect, replicate, and zero padding. The padding size specifies the number of pixels added to the borders of the input feature map.

As shown in Table~\ref{table:padding}, the padding size has little impact on the performance of the reverse convolution operator. In contrast, the choice of padding mode plays a more significant role. Among all evaluated modes, circular padding consistently achieves the best results. This observation is consistent with prior findings in deblurring literature, where circular padding is known to better handle boundary conditions and reduce artifacts. Based on this, we adopt circular padding as the default setting in our operator.

\begin{table}[hpbt]
\caption{The average PSNR(dB) results of Converse-DnCNN with different kernel padding modes and padding sizes.}
\raggedright
\begin{minipage}{0.1\textwidth}
\centering
\footnotesize
\begin{tabular}{>{\centering\arraybackslash}p{1.2cm}>{\centering\arraybackslash}p{0.8cm}>{\centering\arraybackslash}p{0.8cm}} 
\toprule
Padding  & \multicolumn{2}{c}{Datasets} \\
\cmidrule{2-3} 
Modes  & Set12 & BSD68\\
\midrule 
Zero & 30.63 & 29.33 \\
\midrule 
Reflect & 30.65 & 29.34 \\
\midrule 
Replicate & 30.66 & 29.35 \\
\midrule 
\rowcolor{lightblue}
Circular & 30.70 & 29.36 \\
\bottomrule
\end{tabular}
\label{table_kernel_1}
\end{minipage}
\hspace{2.34cm}
\begin{minipage}{0.1\textwidth}
\raggedright
\footnotesize
\begin{tabular}{>{\centering\arraybackslash}p{1.2cm}>{\centering\arraybackslash}p{0.8cm}>{\centering\arraybackslash}p{0.8cm}} 
\toprule
Padding  &  \multicolumn{2}{c}{Datasets} \\
\cmidrule{2-3} 
Sizes& Set12 & BSD68\\
\midrule 
$0\times0$ &  30.67 & 29.34 \\
\midrule 
$2\times2$ & 30.68 & 29.35 \\
\midrule 
\rowcolor{lightblue}
$4\times4$ & 30.70 & 29.36 \\
\midrule 
$5\times5$ & 30.67 & 29.34 \\
\bottomrule
\end{tabular}
\label{table_kernel_2}
\end{minipage}
\label{table:padding}
\end{table}

\vspace{0.2cm}
\noindent \textbf{Regularization Parameter $\lambda$:} The regularization parameter $\lambda$ plays a key role in stabilizing the solution to the ill-posed inverse problem inherent in the Converse2D operation.
Since Converse2D is applied in a channel-wise manner, it is natural to assign separate regularization parameters for each channel to account for variations in feature statistics. To this end, we parameterize $\lambda$ using a learnable transformation:
\begin{equation}
\lambda = \text{Sigmoid}(b - 9.0) + \epsilon
\end{equation}
where $b$ is a learnable scalar and $\epsilon = 1 \times 10^{-5}$ is a small constant added for numerical stability. All $b$ values are initialized to zero, allowing $\lambda$ to be adaptively learned during training. The subtraction of 9.0 shifts the Sigmoid function to produce small initial values for $\lambda$, which helps encourage strong data fidelity at the beginning of training. This formulation enables the model to learn channel-specific regularization strengths in a stable and data-driven manner.

\vspace{0.2cm}
\noindent \textbf{Initial Estimation $\mathbf{X}_0$:}
To evaluate the effect of $\mathbf{X}_0$ initialization on denoising and super-resolution, we compare two strategies from Eq.~\eqref{eq:5}: zero initialization and $\text{Interp}(\mathbf{Y}, s)$, which denotes an upsampled version of $\mathbf{Y}$ by a scale factor $s$. Table~\ref{table:z0} shows clear performance differences between the two strategies across both tasks.
For denoising, initializing with $\text{Interp}(\mathbf{Y}, 1)$ (\ie, $\mathbf{Y}$) leads to significantly higher PSNR, indicating better preservation of high-frequency details critical for noise removal.
In super-resolution, $\text{Interp}(\mathbf{Y}, s)$ similarly yields better results, likely because it provides a closer approximation to the high-resolution target. In contrast, zero initialization results in lower PSNR, potentially due to slower convergence and reduced capacity to recover fine structures.
These results demonstrate that initializing $\mathbf{X}_0$ with $\text{Interp}(\mathbf{Y}, s)$ consistently improves performance by stabilizing optimization and accelerating convergence toward high-quality solutions.

\begin{table}[hpbt]
\caption{The average PSNR(dB) results of Converse-DnCNN and Converse-SRResNet with different initial estimations $\mathbf{X}_0$.}

\begin{minipage}{0.1\textwidth}
\raggedright
\footnotesize
\begin{tabular}{>{\centering\arraybackslash}p{1.2cm} >{\centering\arraybackslash}p{0.5cm} >{\centering\arraybackslash}p{1.05cm}}
\toprule
\multirow{2}{*}[-0.3em]{\vspace{-0.4mm}$\mathbf{X}_0$} &  \multicolumn{2}{c}{Datasets} \\
\cmidrule{2-3} 
& Set12 & BSD68 \\ 
\midrule 
$\mathbf{0}$ & 30.66 & 29.34 \\
\midrule 
\rowcolor{lightblue}
$\mathbf{Y}$ & 30.70 & 29.36 \\
\bottomrule
\end{tabular}
\label{table_kernel_1}
\end{minipage}
\hspace{2.3cm}
\begin{minipage}{0.1\textwidth}
\raggedright
\footnotesize
\begin{tabular}{>{\centering\arraybackslash}p{1.3cm} >{\centering\arraybackslash}p{0.5cm} >{\centering\arraybackslash}p{1.05cm}}
\toprule
\multirow{2}{*}[-0.3em]{\vspace{-0.4mm}$\mathbf{X}_0$} &  \multicolumn{2}{c}{Datasets} \\
\cmidrule{2-3} 
 & Set5 & Urban100 \\ 
\midrule 
$\mathbf{0}$ & 32.22 & 26.24 \\
\midrule 
\rowcolor{lightblue}
$\text{Interp}(\mathbf{Y},s)$ & 32.25 & 26.26 \\
\bottomrule
\end{tabular}
\label{table:z0}
\end{minipage}
\end{table}

%% file: sec/5_experiment.tex
\section{Applications to Image Restoration}

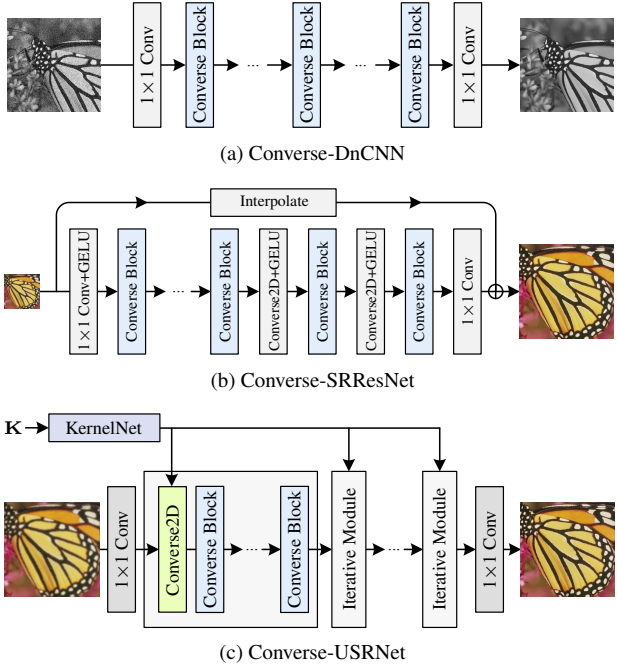
\begin{figure}[!bp]
    \centering
    \begin{subfigure}[b]{0.49\textwidth}
        \centering
        \begin{overpic}[width=0.95\linewidth]{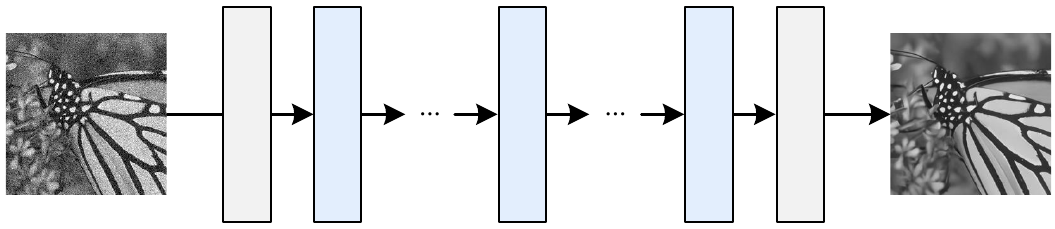}
            \put(22.35,4.26){\rotatebox{90}{\fontsize{7}{7}\selectfont\color{black}{1$\times$1 Conv}}}
            \put(30.93,1.28){\rotatebox{90}{\fontsize{7}{7}\selectfont\color{black}{Converse Block}}}
            \put(48.46,1.28){\rotatebox{90}{\fontsize{7}{7}\selectfont\color{black}{Converse Block}}}
            \put(66.18,1.28){\rotatebox{90}{\fontsize{7}{7}\selectfont\color{black}{Converse Block}}}
            \put(74.89,4.26){\rotatebox{90}{\fontsize{7}{7}\selectfont\color{black}{1$\times$1 Conv}}}
        \end{overpic}
        \vspace{0.05cm}
        \caption{Converse-DnCNN}
        \label{fig:dncnn}
    \end{subfigure}
    \hfill
    \begin{subfigure}[b]{0.49\textwidth}
        \centering
        \vspace{0.3cm}
        \begin{overpic}[width=0.95\linewidth]{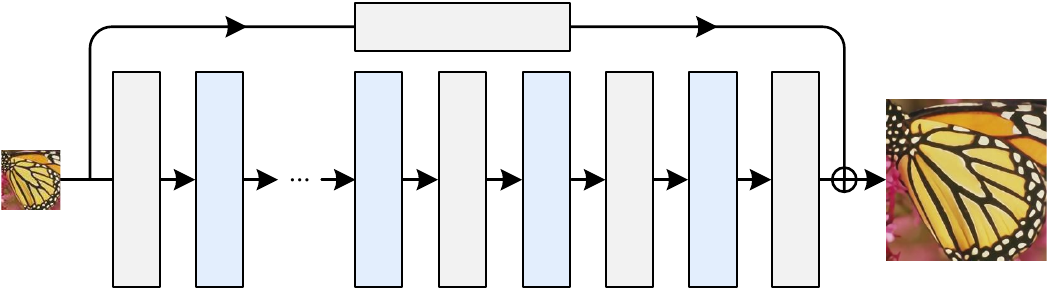}
            \put(12.16,1.15){\rotatebox{90}{\fontsize{6}{8}\selectfont\color{black}{1$\times$1 Conv+GELU}}}
            \put(19.97,2.6){\rotatebox{90}{\fontsize{6}{8}\selectfont\color{black}{Converse Block}}}
            \put(35.25,2.6){\rotatebox{90}{\fontsize{6}{8}\selectfont\color{black}{Converse Block}}}
            \put(43.33,1.12){\rotatebox{90}{\fontsize{5.6}{8}\selectfont\color{black}{Converse2D+GELU}}}
            \put(51.19,2.6){\rotatebox{90}{\fontsize{6}{8}\selectfont\color{black}{Converse Block}}}
            \put(59.2,1.12){\rotatebox{90}{\fontsize{5.6}{8}\selectfont\color{black}{Converse2D+GELU}}}
            \put(66.95,2.6){\rotatebox{90}{\fontsize{6}{8}\selectfont\color{black}{Converse Block}}}
            \put(74.88,5.2){\rotatebox{90}{\fontsize{6}{8}\selectfont\color{black}{1$\times$1 Conv}}}
            \put(38.61,24.8){\fontsize{6}{8}\selectfont\color{black}{Interpolate}}
        \end{overpic}
        \vspace{0.05cm}
        \caption{Converse-SRResNet}
        \label{fig:srresnet}
    \end{subfigure}
    \hfill
    \begin{subfigure}[b]{0.49\textwidth}
        \centering
        \vspace{0.25cm}
        \begin{overpic}[width=0.95\linewidth]{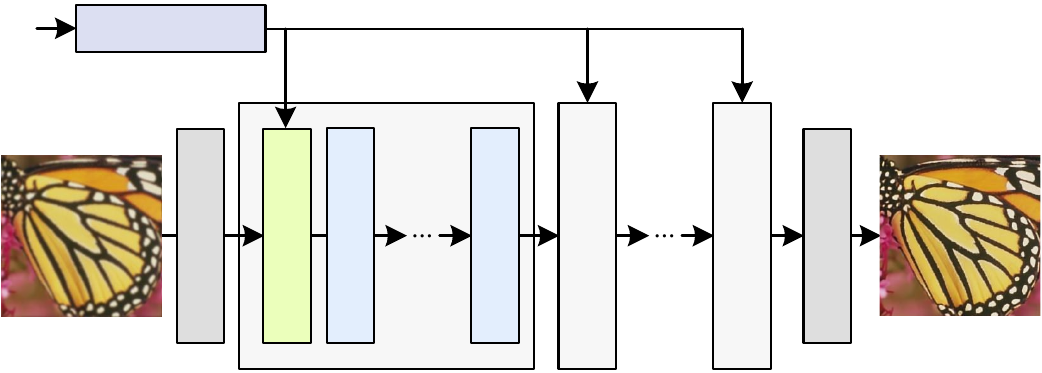}
            \put(18.16,6.65){\rotatebox{90}{\fontsize{7}{8}\selectfont\color{black}{1$\times$1 Conv}}}
            \put(26.5,5.61){\rotatebox{90}{\fontsize{7}{8}\selectfont\color{black}{Converse2D}}}
            \put(32.63,3.6){\rotatebox{90}{\fontsize{7}{8}\selectfont\color{black}{Converse Block}}}
            \put(46.42,3.6){\rotatebox{90}{\fontsize{7}{8}\selectfont\color{black}{Converse Block}}}
            \put(55.4,3.18){\rotatebox{90}{\fontsize{7}{8}\selectfont\color{black}{Iterative Module}}}
            \put(70.16,3.18){\rotatebox{90}{\fontsize{7}{8}\selectfont\color{black}{Iterative Module}}}
            \put(78.35,6.65){\rotatebox{90}{\fontsize{7}{8}\selectfont\color{black}{1$\times$1 Conv}}}
            \put(0,32.05){\fontsize{7}{8}\selectfont\color{black}{$\mathbf{K}$}}
            \put(10.05,32.1){\fontsize{7}{8}\selectfont\color{black}{KernelNet}}
        \end{overpic}
        \vspace{0.05cm}
        \caption{Converse-USRNet}
        \label{fig:usrnet}
    \end{subfigure}
    \caption{Architectures of the proposed converse-based networks for denoising, super-resolution, and deblurring.}
    \label{fig:all_converse_archs}
\end{figure}

To evaluate the effectiveness of the proposed Converse2D operator and Converse Block, we consider two classical image restoration tasks: Gaussian denoising and super-resolution.
Following the architecture designs of DnCNN and SRResNet, we replace their core building blocks with the proposed Converse2D operator and converse block, resulting in two variants: Converse-DnCNN and Converse-SRResNet.
The architectures of these models are shown in Fig.~\ref{fig:dncnn} and Fig.~\ref{fig:srresnet}, containing 20 and 16 converse blocks, respectively.

Second, to demonstrate the applicability of our operator in kernel-conditioned scenarios, we consider non-blind deblurring.
We modify USRNet by replacing its data module with our Converse2D operator and substituting the ResUNet denoiser with a ConverseNet consisting of 7 Converse Blocks.
Since Converse2D supports arbitrary channel dimensions, we first project the input image to a 64-channel feature map using a $1 \times 1$ convolution, and map it back to 3 channels at the output.
We also design KernelNet, which maps the blurred kernel to a 64-dimensional embedding via three fully connected layers with GELU activation, followed by a $1 \times 1$ convolution.
The resulting kernel embedding is directly used to parameterize Converse2D, with $\lambda$ as the only learnable parameter.
The overall architecture of Converse-USRNet is illustrated in Fig.~\ref{fig:usrnet}.

For the Converse2D operator, we use a kernel size of $5 \times 5$, circular padding of size 4, and initialize $\mathbf{X}_0$ with $\text{Interp}(\mathbf{Y}, s)$ as the default setting.
To focus on validating the effectiveness of the proposed operator rather than achieving state-of-the-art performance, we adopt simplified experimental settings across all tasks.
For denoising, we use Gaussian noise with a standard deviation of $\sigma = 25$. For super-resolution, we consider $\times4$ upscaling from bicubically downsampled images. For deblurring, we employ both motion and Gaussian blur kernels with a kernel size of 7 and additive noise of standard deviation 2.55.
We follow the training settings in~\cite{zhang2021plug}, including dataset selection, loss function, and learning rate.

\subsection{Image Denoising}

\begin{table}[tbp!]
\raggedright
\caption{{The number of parameters and average PSNR(dB) results of different models for Gaussian denoising with noise level 25 on Set12 and BSD68 datasets.}}
\renewcommand\arraystretch{1.00}
\resizebox{0.48\textwidth}{!}
{
\begin{tabular}{>{\centering\arraybackslash}p{3.5cm}>{\centering\arraybackslash}p{2.2cm}>{\centering\arraybackslash}p{0.7cm}>{\centering\arraybackslash}p{0.9cm}}
\toprule
\multirow{2}{*}[-0.18em]{Models}   & \multirow{2}{*}[-0.18em]{\# Parameters} &\multicolumn{2}{c}{Datasets} \\
\cmidrule{3-4} 
\begin{tabular}[c]{@{}l@{}} \end{tabular} & & Set12 &  BSD68 \\ 
\midrule
DnCNN~\cite{dncnn}&557,057 & 30.43 & 29.23 \\
\midrule
Conv-DnCNN&  734,913& 30.64 & 29.30  \\
ConvT-DnCNN& 734,913& 30.61 & 29.29  \\
\rowcolor{lightblue}
Converse-DnCNN& 734,913& 30.70 & 29.36 \\ 
 \bottomrule
\end{tabular}
}
\label{table:denoising}
\end{table}

\begin{figure*}[!h]
\centering
\subfloat[\scriptsize Noisy image]
{\includegraphics[width=0.19\textwidth]{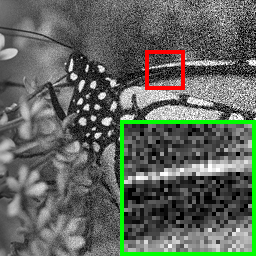}}
\hspace{1mm}
\subfloat[\scriptsize DnCNN/23.49dB]
{\includegraphics[width=0.19\textwidth]{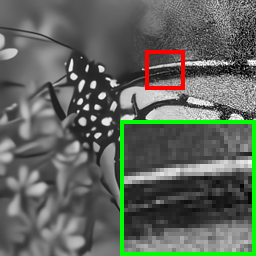}}
\hspace{1mm}
\subfloat[\scriptsize Conv-DnCNN/23.40dB]
{\includegraphics[width=0.19\textwidth]{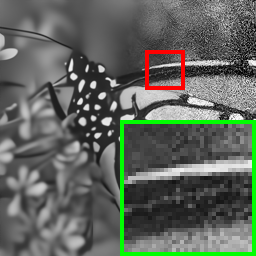}}
\hspace{1mm}
\subfloat[\scriptsize ConvT-DnCNN/23.24dB]
{\includegraphics[width=0.19\textwidth]{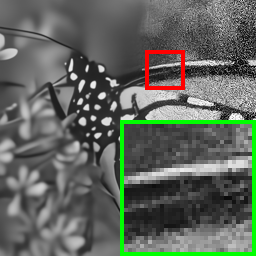}}
\hspace{1mm}
\subfloat[\scriptsize Converse-DnCNN/23.53dB]
{\includegraphics[width=0.19\textwidth]{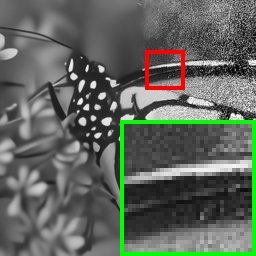}}
\vspace{-0.1cm}
\caption{Denoising results of different methods on the image ``butterfly'' from Set12 with noise level uniformly ranging from 0 on the left to 50 on the right.}
\label{figure:0-50}
\end{figure*}

\begin{figure*}[!htbp]
\centering
\subfloat[\scriptsize Low-resolution image]
{\includegraphics[width=0.19\textwidth]{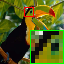}}
\hspace{1mm}
\subfloat[\scriptsize SRResNet/32.90dB]
{\includegraphics[width=0.19\textwidth]{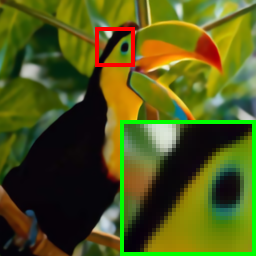}}
\hspace{1mm}
\subfloat[\scriptsize Conv-SRResNet/32.79dB]
{\includegraphics[width=0.19\textwidth]{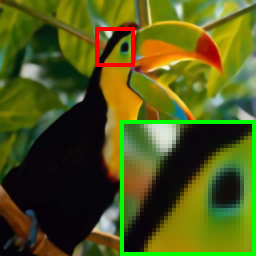}}
\hspace{1mm}
\subfloat[\scriptsize ConvT-SRResNet/32.59dB]
{\includegraphics[width=0.19\textwidth]{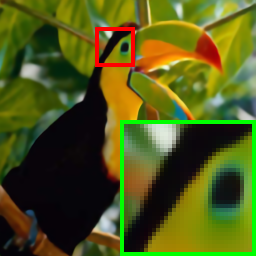}}
\hspace{1mm}
\subfloat[\scriptsize Converse-SRResNet/32.89dB]
{\includegraphics[width=0.19\textwidth]{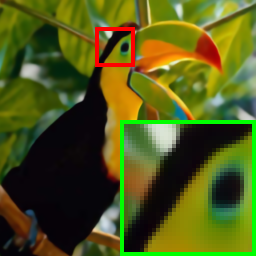}}
\vspace{-0.1cm}
\caption{Single image super-resolution results of different models on the image ``bird'' from Set5 with an upscaling factor of 4.}
\label{figure:sr_results}
\end{figure*}

\begin{figure*}[!h]
\centering
\subfloat[\scriptsize Blurry image]
{\includegraphics[width=0.19\textwidth]{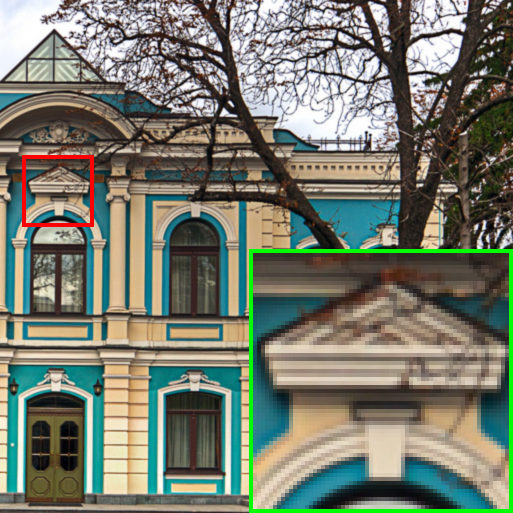}}
\hspace{1mm}
\subfloat[\scriptsize ConvNet/20.93dB]
{\includegraphics[width=0.19\textwidth]{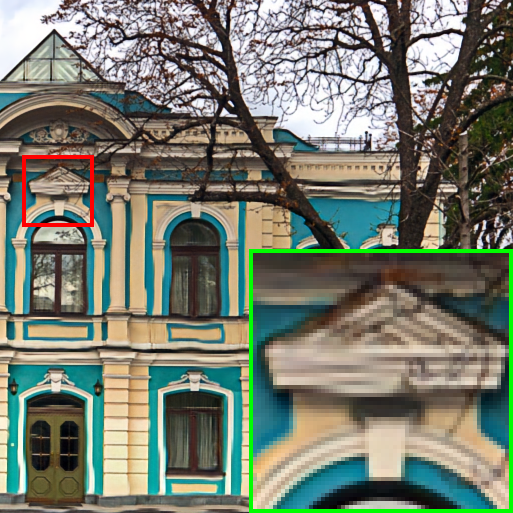}}
\hspace{1mm}
\subfloat[\scriptsize ConverseNet/22.73dB]
{\includegraphics[width=0.19\textwidth]{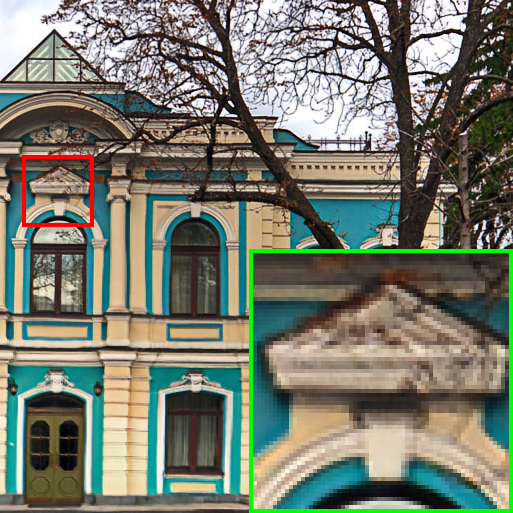}}
\hspace{1mm}
\subfloat[\scriptsize USRNet/34.20dB]
{\includegraphics[width=0.19\textwidth]{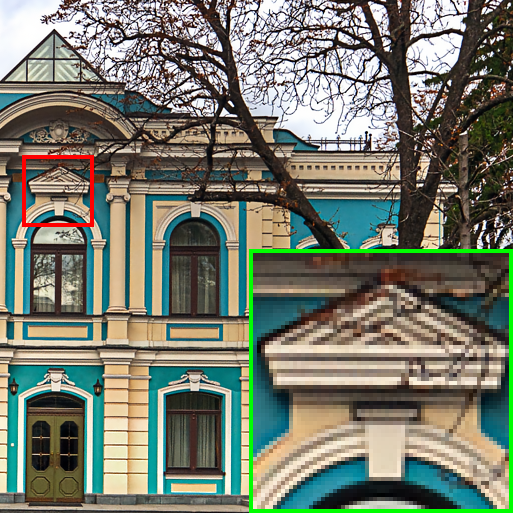}}
\hspace{1mm}
\subfloat[\scriptsize Converse-USRNet/34.27dB]
{\includegraphics[width=0.19\textwidth]{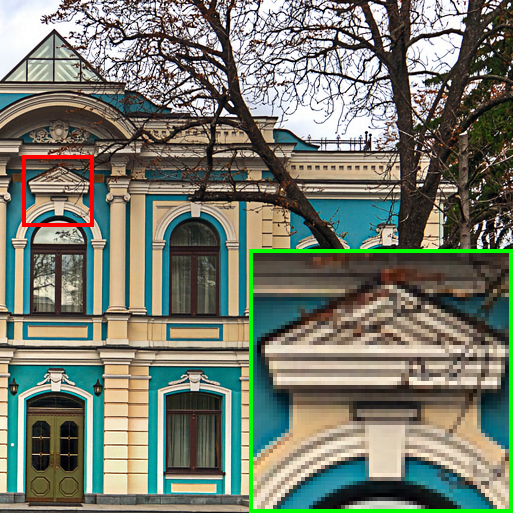}}
\vspace{-0.1cm}
\caption{Deblurring results of different methods on the image ``building'' from Urban100 with noise level 2.55.}
\label{figure:converse-usrnet}
\vspace{-0.25cm}
\end{figure*}

In our denoising experiments, we evaluate the proposed Converse-DnCNN by comparing it with the original DnCNN and two modified variants: Conv-DnCNN and ConvT-DnCNN, where the Converse2D operator is replaced by depthwise convolution and depthwise transposed convolution, respectively.
Table~\ref{table:denoising} reports the number of parameters and average PSNR values on the Set12 and BSD68 datasets~\cite{roth2009fields,dncnn,martin2001database} for Gaussian noise with $\sigma = 25$.
Although Converse-DnCNN includes 20 Converse Blocks and thus has more parameters than the original DnCNN, it achieves a substantial PSNR improvement.
DnCNN is included as a baseline for reference, while the main comparison focuses on the effectiveness of the proposed operator.
As shown in Table~\ref{table:denoising}, Converse-DnCNN consistently outperforms both Conv-DnCNN and ConvT-DnCNN, demonstrating the effectiveness of the Converse2D operator for image denoising.

Fig.~\ref{figure:0-50} shows the visual results comparison between different models on a noisy image ``butterfly'' from Set12 with noise level uniformly ranging from 0 on the left and 50 on the right. We emphasize here that deconvolution algorithms tend to produce border artifacts due to the lack of effective boundary constraints. This limitation stems from the inherent properties of deconvolution. To address this issue, various techniques, such as padding strategies and regularization terms, are often commonly employed. However, these methods can only partially solve the problem. In contrast, our Converse-DnCNN model is trained end-to-end and does not introduce border artifacts in the denoised results.

\subsection{Image Super-Resolution}

\begin{table}[htbp]
\caption{{The average PSNR(dB) results of different variants of SRResNet for $\times4$ super-resolution.}}
\renewcommand\arraystretch{1.00}
\resizebox{0.48\textwidth}{!}
{
\begin{tabular}{ccccc}
\toprule
\multirow{2}{*}[-0.18em]{Models} &  \multicolumn{4}{c}{Datasets} \\
\cmidrule{2-5}
\begin{tabular}[c]{@{}c@{}}  \end{tabular} & Set5 &  Set14 & BSD100 & Urban100\\ \midrule
SRResNet~\cite{srresnet} & 32.21 & 28.60 & 27.59 & 26.09 \\
\midrule
Conv-SRResNet & 32.23 & 28.73 & 27.62 & 26.24 \\
ConvT-SRResNet & 32.09 & 28.61 & 27.56 & 25.89 \\
\rowcolor{lightblue}
Converse-SRResNet & 32.25 & 28.72 & 27.62 & 26.24 \\
 \bottomrule
\end{tabular}
}
\label{table:sr}
\end{table}

\begin{table}[htbp]
\caption{The average PSNR(dB) results of different variants of SRResNet with different upscaling methods for $\times4$ super-resolution.}
\renewcommand\arraystretch{1.00}
\resizebox{0.48\textwidth}{!}
{
\begin{tabular}{ccccc}
\toprule
Upscaling  &  \multicolumn{4}{c}{Datasets} \\
\cmidrule{2-5} 
\begin{tabular}[c]{@{}c@{}} Methods \end{tabular} & Set5 &  Set14 & BSD100 & Urban100\\ \midrule
Pixelshuffle (Default) & 32.26 & 28.74 & 27.65 & 26.33 \\
Nearest Interpolation & 32.26 & 28.77 & 27.65 & 26.36 \\
ConvT & 32.24 & 28.74 & 27.64 & 26.33 \\
\rowcolor{lightblue}
Converse2D & 32.26 & 28.78 & 27.66 & 26.36 \\
\bottomrule
\end{tabular}
}
\label{table:upsample}
\end{table}

In our super-resolution experiments, we evaluate the proposed Converse-SRResNet by comparing it with the original SRResNet~\cite{srresnet} and two modified versions of Converse-SRResNet, named Conv-SRResNet and ConvT-SRResNet. In these two modified versions, the Converse2D operator is replaced with depthwise convolution and depthwise transposed convolution, respectively. Table~\ref{table:sr} provides a comparison of PSNR results for each model on different datasets~\cite{timofte2014a+,bevilacqua2012low}.
We can see that our Converse-SRResNet achieves performance comparable to the other models. This indicates that Converse2D can effectively replace its counterpart operators without reducing performance.
Additionally, we conduct experiments by replacing the original upsampling module in SRResNet with different upsampling methods, including nearest-neighbor interpolation, transposed convolution, and Converse2D. Table~\ref{table:upsample} shows that our proposed Converse2D operator performs on par with these other upsampling methods. This demonstrates the effectiveness of Converse2D as an upsampling module for super-resolution tasks. Fig.~\ref{figure:sr_results} compares visual outputs across models, demonstrating that our Reverse-SRResNet achieves comparable reconstruction quality.

\subsection{Image Deblurring}

In our deblurring experiments, we evaluate the proposed Converse-USRNet in a non-blind setting by comparing it with Conv-USRNet, where the data module operates on three-channel inputs. To further assess the impact of the proposed converse operator, we compare against ConvNet and ConverseNet, both implemented by removing the KernelNet and operating in a blind deblurring setting without access to kernel information.
Table~\ref{table:deblur} reports average PSNR results, where blurred images are synthesized by convolving clean images with a $7 \times 7$ motion blur kernel and adding Gaussian noise with a standard deviation of 2.55.
Note that $7 \times 7$ isotropic Gaussian blur kernels are also used during training for both blind and non-blind models.

\begin{table}[!htbp]\fontsize{6.85}{10.5}\selectfont
\caption{Average PSNR(dB) of different methods for image deblurring under various blur kernels on the BSD100 and Urban100 datasets.} 
\center
\begin{tabular}{>{\centering\arraybackslash}p{1.8cm}|>{\centering\arraybackslash}p{0.5cm}|>{\centering\arraybackslash}p{0.6cm}|>{\centering\arraybackslash}m{0.6cm}|>{\centering\arraybackslash}m{0.6cm}|>{\centering\arraybackslash}m{0.6cm}|>{\centering\arraybackslash}m{0.6cm}}
  \hline

  \multirow{4}{*}{Methods} & \multirow{4}{*}{Blind} &\multicolumn{5}{c}{Blur Kernels} \\ \cline{3-7}

  & & \multirow{3}{*}{\vspace{-0.115cm}\hspace{-0.035cm}\includegraphics[width=0.0385\textwidth]{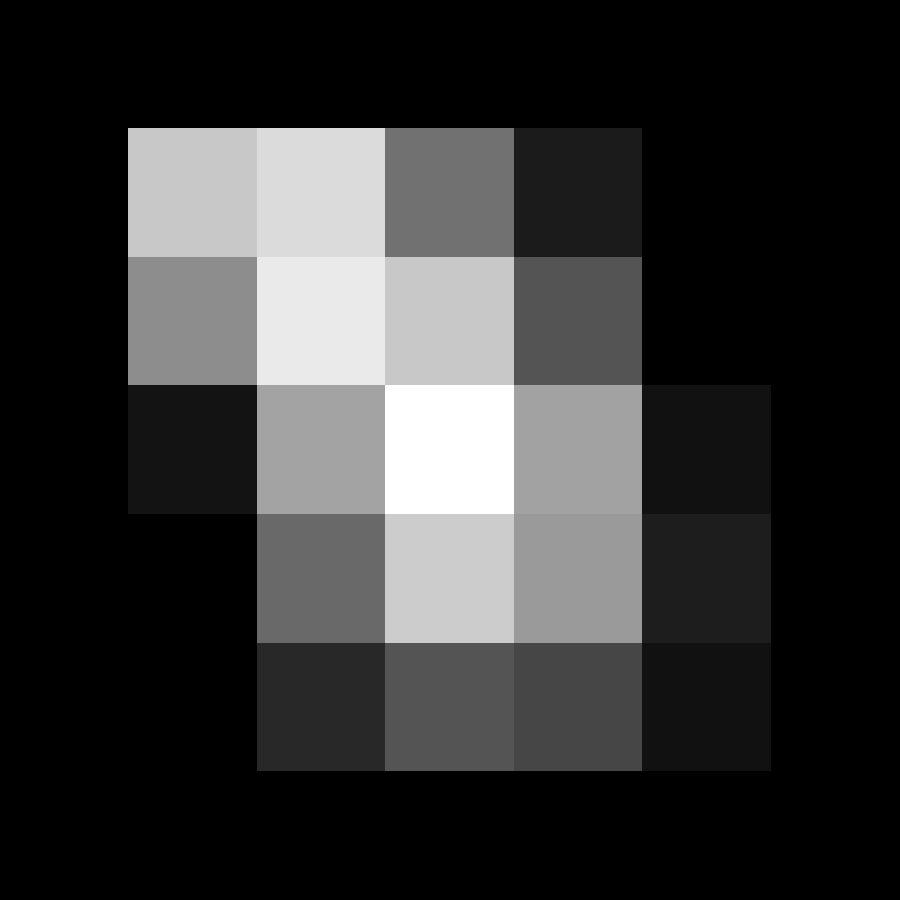}} & \multirow{3}{*}{\vspace{-0.115cm}\hspace{-0.035cm}\includegraphics[width=0.0385\textwidth]{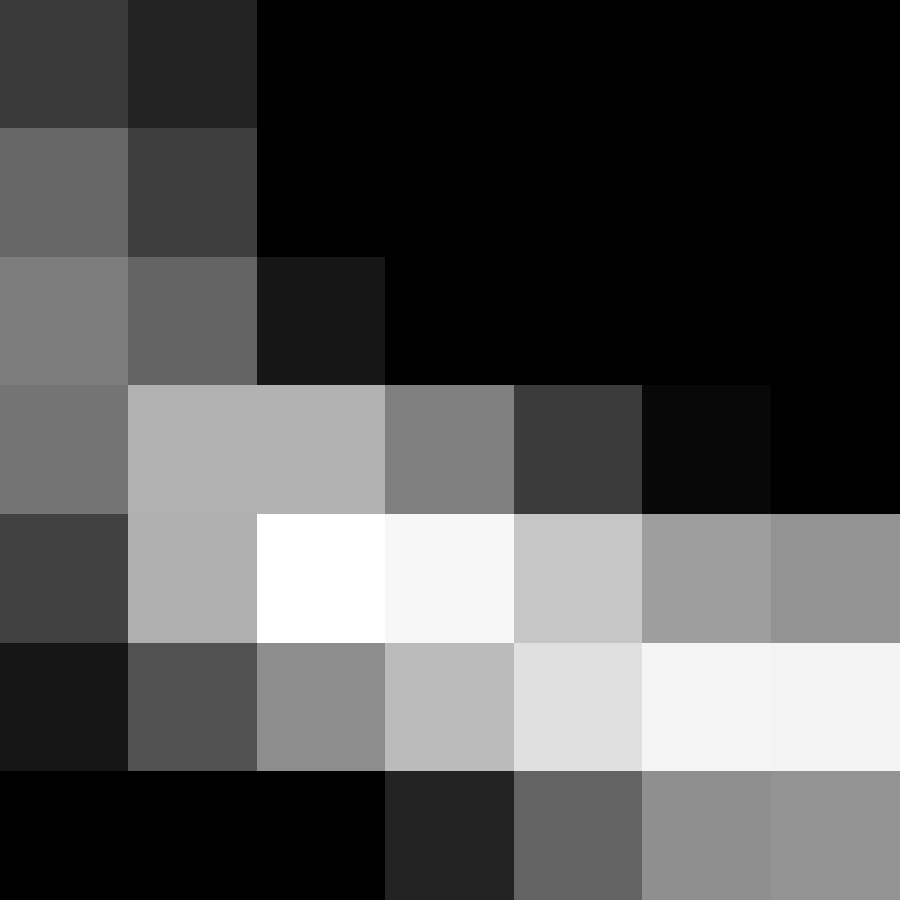}} & \multirow{3}{*}{\vspace{-0.115cm}\hspace{-0.035cm}\includegraphics[width=0.0385\textwidth]{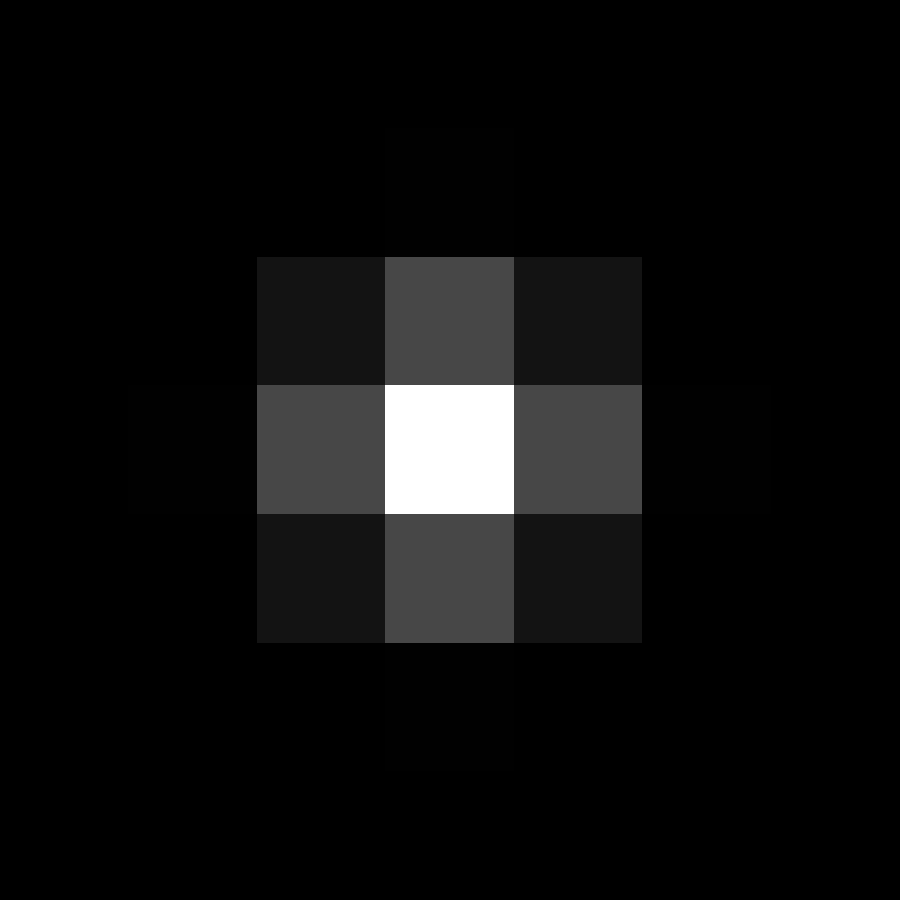}} & \multirow{3}{*}
  {\vspace{-0.115cm}\hspace{-0.035cm}\includegraphics[width=0.0385\textwidth]{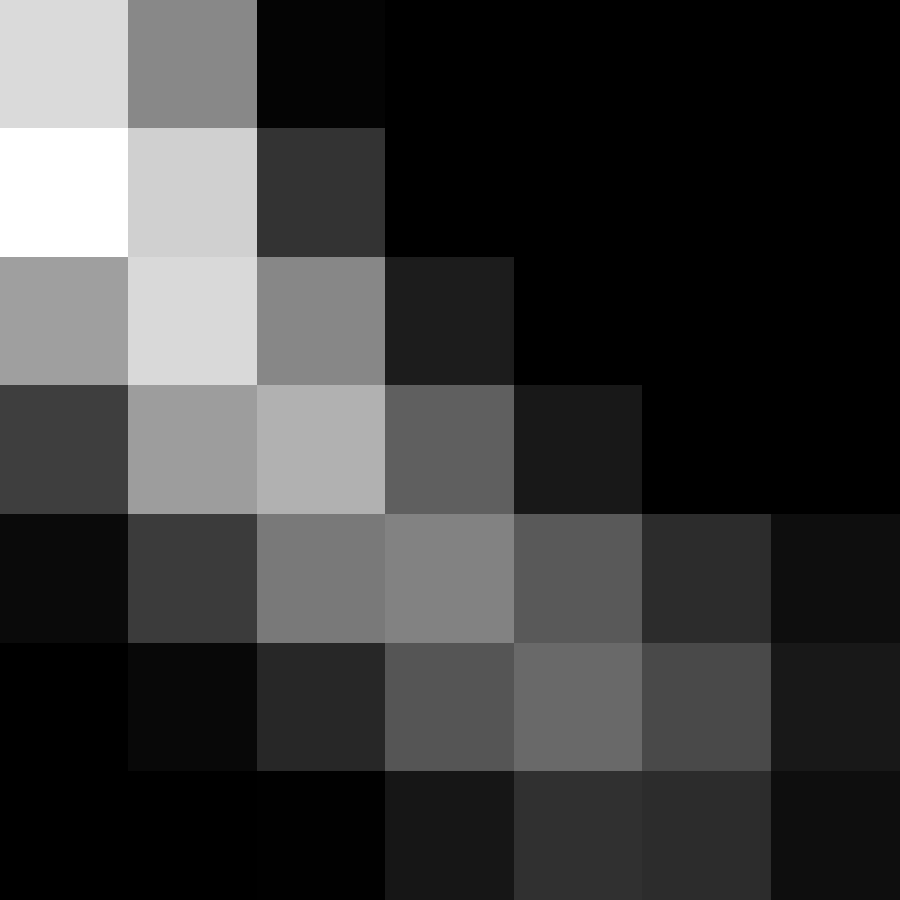}} & \multirow{3}{*}
  {\vspace{-0.115cm}\hspace{-0.035cm}\includegraphics[width=0.0385\textwidth]{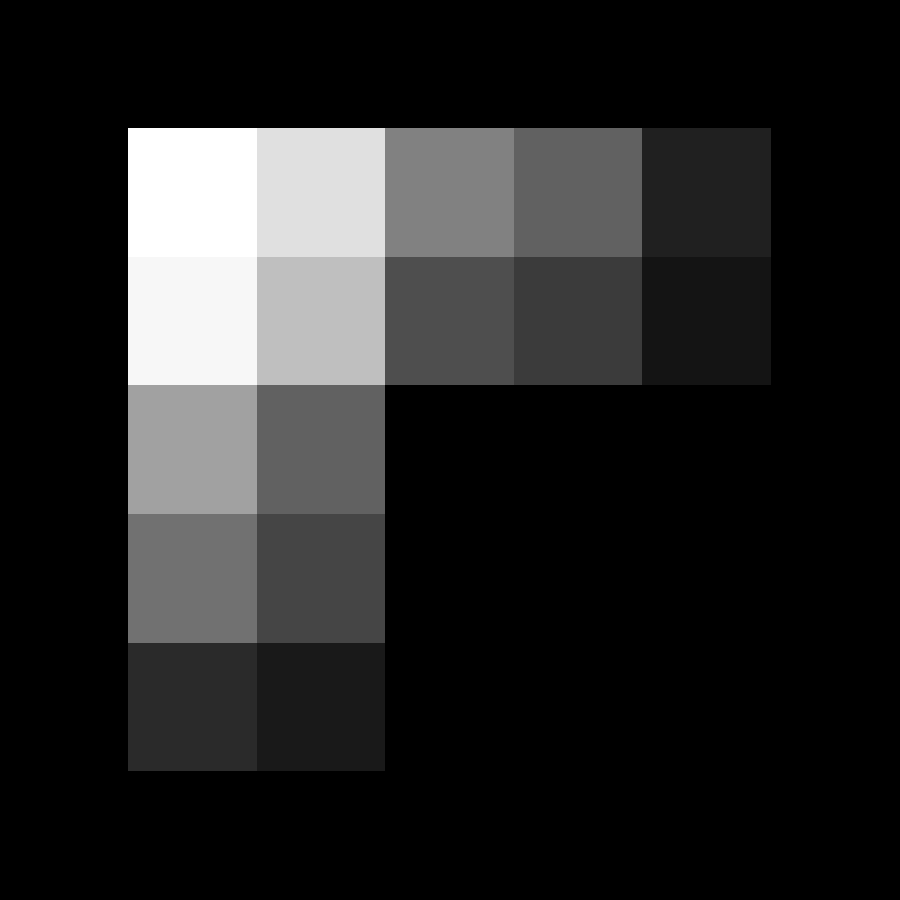}}\\
    &  &  &  &  &  &  \\
    &  &  &  &  &  &  \\ \hline\hline

Datasets & & \multicolumn{5}{c}{BSD100}  \\
\hline

ConvNet & \ding{51} & 25.26 & 22.96 & 27.16 & 26.91 & 23.98\\

ConverseNet & \ding{51} & 27.69 & 24.43 & 31.28 & 27.64 & 27.31\\

Conv-USRNet & \ding{55} & 32.18 & 31.03 & 37.80 & 31.82 & 34.33\\

\rowcolor{lightblue}

Converse-USRNet & \ding{55} & 32.46 & 31.20 & 38.40 & 32.21 & 34.62\\

\hline
\hline

Datasets & & \multicolumn{5}{c}{Urban100}  \\
\hline

ConvNet & \ding{51} & 22.27 & 20.59 & 23.43 & 23.92 & 21.12\\

ConverseNet & \ding{51} & 24.24 & 21.52 & 23.16 & 24.56 & 22.56\\ 

Conv-USRNet & \ding{55} & 31.48 & 30.19 & 36.80 & 31.23 & 33.63\\

\rowcolor{lightblue}

Converse-USRNet & \ding{55} & 31.96 & 30.59 & 37.48 & 31.95 & 33.99\\
\hline
\end{tabular}
\label{table:deblur}
\end{table}

We can have two key observations from Table~\ref{table:deblur}.
First, in the non-blind setting, Converse-USRNet outperforms its counterpart Conv-USRNet. The performance gain over blind models can be largely attributed to kernel conditioning and multi-channel processing enabled by the Converse2D framework.
While Converse-USRNet and USRNet share similar overall architectures, the main distinction lies in their design: USRNet performs deconvolution and denoising directly in the image domain, whereas Converse-USRNet operates in the feature domain. This design offers greater flexibility and learning capacity.
Moreover, the proposed reverse convolution operator enables Eq.~\eqref{eq:solution} to serve as a general module that jointly incorporates kernel information and captures spatial dependencies.

Second, in the blind setting, ConverseNet consistently outperforms ConvNet. This improvement is primarily attributed to the inherent deconvolution capability of the Converse2D operator, even without explicit kernel input.
As shown in Fig.~\ref{figure:converse-usrnet}, ConvNet introduces geometric distortions; for example, straight lines such as walls may appear bent or curved. In contrast, ConverseNet mitigates these artifacts and produces sharper, more structurally consistent results.
By contrast, both Conv-USRNet and Converse-USRNet, operating in the non-blind setting with access to kernel information, generate well-aligned and visually sharp outputs with accurately preserved straight structures.

%% file: sec/6_conclusion.tex
\section{Conclusion}

In this paper, we proposed a novel depthwise reverse convolution operator that serves as an explicit inversion of depthwise convolution by solving a regularized least-squares problem in closed form. Unlike conventional transposed convolution, the proposed operator provides mathematically grounded inversion and enables stable, efficient integration into deep networks.
To enhance its applicability, we further introduced a reverse convolution block that combines the operator with layer normalization, $1 \times 1$ convolutions, and GELU activation. This design decouples spatial modeling and channel interaction, reflecting the modular structure commonly used in Transformer-based architectures.
We applied the proposed block to three representative image restoration tasks: Gaussian denoising, super-resolution, and deblurring, by constructing Converse-DnCNN, Converse-SRResNet, and Converse-USRNet. Extensive experiments demonstrated that the proposed operator achieves competitive or superior performance compared to conventional convolution and transposed convolution, validating its effectiveness as a general-purpose module.

This work provides a principled alternative to transposed convolution and opens new directions for designing learnable inverse operators in deep architectures. Future work includes extending this operator to large-scale vision models and generative tasks such as image synthesis and restoration under complex degradations.

\vspace{0.3cm}
\noindent\textbf{Acknowledgments:}
This work was supported by Suzhou Key Technologies Project (Grant No. SYG2024136), Natural Science Foundation of China (Grant No. 62406135), and Natural Science Foundation of Jiangsu Province (Grant No. BK20241198).

%% file: main.bbl
\begin{thebibliography}{50}
\providecommand{\natexlab}[1]{#1}
\providecommand{\url}[1]{\texttt{#1}}
\expandafter\ifx\csname urlstyle\endcsname\relax
  \providecommand{\doi}[1]{doi: #1}\else
  \providecommand{\doi}{doi: \begingroup \urlstyle{rm}\Url}\fi

\bibitem[Behrmann et~al.(2019)Behrmann, Grathwohl, Chen, Duvenaud, and Jacobsen]{DBLP:conf/icml/BehrmannGCDJ19}
Jens Behrmann, Will Grathwohl, Ricky T.~Q. Chen, David Duvenaud, and J{\"{o}}rn{-}Henrik Jacobsen.
\newblock Invertible residual networks.
\newblock In \emph{{ICML}}, pages 573--582. {PMLR}, 2019.

\bibitem[Bevilacqua et~al.(2012)Bevilacqua, Roumy, Guillemot, and Alberi-Morel]{bevilacqua2012low}
Marco Bevilacqua, Aline Roumy, Christine Guillemot, and Marie~Line Alberi-Morel.
\newblock \emph{Low-complexity single-image super-resolution based on nonnegative neighbor embedding}.
\newblock BMVA press, 2012.

\bibitem[Chen et~al.(2024)Chen, Gu, Zheng, and Fu]{chen2024frequency}
Linwei Chen, Lin Gu, Dezhi Zheng, and Ying Fu.
\newblock Frequency-adaptive dilated convolution for semantic segmentation.
\newblock In \emph{CVPR}, pages 3414--3425, 2024.

\bibitem[Chen(2015)]{chen2014semantic}
Liang-Chieh Chen.
\newblock Semantic image segmentation with deep convolutional nets and fully connected crfs.
\newblock In \emph{ICLR}, 2015.

\bibitem[Chen et~al.(2019)Chen, Behrmann, Duvenaud, and Jacobsen]{DBLP:conf/nips/ChenBDJ19}
Tian~Qi Chen, Jens Behrmann, David Duvenaud, and J{\"{o}}rn{-}Henrik Jacobsen.
\newblock Residual flows for invertible generative modeling.
\newblock In \emph{NeurIPS}, pages 9913--9923, 2019.

\bibitem[Chollet(2017)]{xception}
Fran{\c{c}}ois Chollet.
\newblock Xception: Deep learning with depthwise separable convolutions.
\newblock In \emph{CVPR}, pages 1251--1258, 2017.

\bibitem[Dinh et~al.(2015)Dinh, Krueger, and Bengio]{dinh2014nice}
Laurent Dinh, David Krueger, and Yoshua Bengio.
\newblock Nice: Non-linear independent components estimation.
\newblock In \emph{ICLR}, 2015.

\bibitem[Dinh et~al.(2017)Dinh, Sohl-Dickstein, and Bengio]{dinh2016density}
Laurent Dinh, Jascha Sohl-Dickstein, and Samy Bengio.
\newblock Density estimation using real nvp.
\newblock In \emph{ICLR}, 2017.

\bibitem[Dong et~al.(2020)Dong, Roth, and Schiele]{dong2020deep}
Jiangxin Dong, Stefan Roth, and Bernt Schiele.
\newblock Deep wiener deconvolution: Wiener meets deep learning for image deblurring.
\newblock \emph{NeurIPS}, 33:\penalty0 1048--1059, 2020.

\bibitem[Gao(2023)]{gao2023rethinking}
Roland Gao.
\newblock Rethinking dilated convolution for real-time semantic segmentation.
\newblock In \emph{CVPR}, pages 4675--4684, 2023.

\bibitem[Grathwohl et~al.(2019)Grathwohl, Chen, Bettencourt, Sutskever, and Duvenaud]{DBLP:conf/iclr/GrathwohlCBSD19}
Will Grathwohl, Ricky T.~Q. Chen, Jesse Bettencourt, Ilya Sutskever, and David Duvenaud.
\newblock {FFJORD:} free-form continuous dynamics for scalable reversible generative models.
\newblock In \emph{{ICLR}}, 2019.

\bibitem[Hendrycks and Gimpel(2016)]{hendrycks2016gaussian}
Dan Hendrycks and Kevin Gimpel.
\newblock Gaussian error linear units (gelus).
\newblock \emph{arXiv}, 2016.

\bibitem[Howard(2017)]{mobilenets}
Andrew~G Howard.
\newblock Mobilenets: Efficient convolutional neural networks for mobile vision applications.
\newblock \emph{arXiv}, 2017.

\bibitem[Hua et~al.(2018)Hua, Tran, and Yeung]{hua2018pointwise}
Binh-Son Hua, Minh-Khoi Tran, and Sai-Kit Yeung.
\newblock Pointwise convolutional neural networks.
\newblock In \emph{CVPR}, pages 984--993, 2018.

\bibitem[Huang et~al.(2018)Huang, Liu, Van~der Maaten, and Weinberger]{huang2018condensenet}
Gao Huang, Shichen Liu, Laurens Van~der Maaten, and Kilian~Q Weinberger.
\newblock Condensenet: An efficient densenet using learned group convolutions.
\newblock In \emph{CVPR}, pages 2752--2761, 2018.

\bibitem[Kamilov et~al.(2017)Kamilov, Mansour, and Wohlberg]{kamilov2017plug}
Ulugbek~S Kamilov, Hassan Mansour, and Brendt Wohlberg.
\newblock A plug-and-play priors approach for solving nonlinear imaging inverse problems.
\newblock \emph{IEEE SPL}, 24\penalty0 (12):\penalty0 1872--1876, 2017.

\bibitem[Khalfaoui-Hassani et~al.(2023)Khalfaoui-Hassani, Pellegrini, and Masquelier]{khalfaoui2021dilated}
Ismail Khalfaoui-Hassani, Thomas Pellegrini, and Timoth{\'e}e Masquelier.
\newblock Dilated convolution with learnable spacings.
\newblock In \emph{ICLR}, 2023.

\bibitem[Kingma and Dhariwal(2018)]{kingma2018glow}
Durk~P Kingma and Prafulla Dhariwal.
\newblock Glow: Generative flow with invertible 1x1 convolutions.
\newblock \emph{NeurIPS}, 31, 2018.

\bibitem[Kumar et~al.(2019)Kumar, Babaeizadeh, Erhan, Finn, Levine, Dinh, and Kingma]{kumar2019videoflow}
Manoj Kumar, Mohammad Babaeizadeh, Dumitru Erhan, Chelsea Finn, Sergey Levine, Laurent Dinh, and Durk Kingma.
\newblock Videoflow: A flow-based generative model for video.
\newblock \emph{arXiv}, 2019.

\bibitem[Kundur and Hatzinakos(1996)]{kundur1996blind}
Deepa Kundur and Dimitrios Hatzinakos.
\newblock Blind image deconvolution.
\newblock \emph{IEEE SPM}, 13\penalty0 (3):\penalty0 43--64, 1996.

\bibitem[LeCun et~al.(1998)LeCun, Bottou, Bengio, and Haffner]{lecun1998gradient}
Yann LeCun, L{\'e}on Bottou, Yoshua Bengio, and Patrick Haffner.
\newblock Gradient-based learning applied to document recognition.
\newblock \emph{Proceedings of the IEEE}, 86\penalty0 (11):\penalty0 2278--2324, 1998.

\bibitem[Ledig et~al.(2017)Ledig, Theis, Husz{\'a}r, Caballero, Cunningham, Acosta, Aitken, Tejani, Totz, Wang, et~al.]{srresnet}
Christian Ledig, Lucas Theis, Ferenc Husz{\'a}r, Jose Caballero, Andrew Cunningham, Alejandro Acosta, Andrew Aitken, Alykhan Tejani, Johannes Totz, Zehan Wang, et~al.
\newblock Photo-realistic single image super-resolution using a generative adversarial network.
\newblock In \emph{CVPR}, pages 4681--4690, 2017.

\bibitem[Lee et~al.(2021)Lee, Son, Rim, Cho, and Lee]{lee2021iterative}
Junyong Lee, Hyeongseok Son, Jaesung Rim, Sunghyun Cho, and Seungyong Lee.
\newblock Iterative filter adaptive network for single image defocus deblurring.
\newblock In \emph{CVPR}, pages 2034--2042, 2021.

\bibitem[Lei~Ba et~al.(2016)Lei~Ba, Kiros, and Hinton]{ba2016layer}
Jimmy Lei~Ba, Jamie~Ryan Kiros, and Geoffrey~E Hinton.
\newblock Layer normalization.
\newblock \emph{arXiv}, 2016.

\bibitem[Martin et~al.(2001)Martin, Fowlkes, Tal, and Malik]{martin2001database}
David Martin, Charless Fowlkes, Doron Tal, and Jitendra Malik.
\newblock A database of human segmented natural images and its application to evaluating segmentation algorithms and measuring ecological statistics.
\newblock In \emph{ICCV}, pages 416--423, 2001.

\bibitem[Mou et~al.(2022)Mou, Wang, and Zhang]{mou2022deep}
Chong Mou, Qian Wang, and Jian Zhang.
\newblock Deep generalized unfolding networks for image restoration.
\newblock In \emph{CVPR}, pages 17399--17410, 2022.

\bibitem[Noh et~al.(2015)Noh, Hong, and Han]{noh2015learning}
Hyeonwoo Noh, Seunghoon Hong, and Bohyung Han.
\newblock Learning deconvolution network for semantic segmentation.
\newblock In \emph{ICCV}, pages 1520--1528, 2015.

\bibitem[Odena et~al.(2016)Odena, Dumoulin, and Olah]{odena2016checkerboard}
Augustus Odena, Vincent Dumoulin, and Chris Olah.
\newblock Deconvolution and checkerboard artifacts.
\newblock \emph{Distill}, 1\penalty0 (10):\penalty0 e3, 2016.

\bibitem[Radford(2016)]{radford2015unsupervised}
Alec Radford.
\newblock Unsupervised representation learning with deep convolutional generative adversarial networks.
\newblock In \emph{ICLR}, 2016.

\bibitem[Ronneberger et~al.(2015)Ronneberger, Fischer, and Brox]{ronneberger2015unet}
Olaf Ronneberger, Philipp Fischer, and Thomas Brox.
\newblock U-net: Convolutional networks for biomedical image segmentation.
\newblock In \emph{MICCAI}, pages 234--241. Springer, 2015.

\bibitem[Roth and Black(2009)]{roth2009fields}
Stefan Roth and Michael~J Black.
\newblock Fields of experts.
\newblock \emph{IJCV}, 82\penalty0 (2):\penalty0 205--229, 2009.

\bibitem[Schuler et~al.(2013)Schuler, Christopher~Burger, Harmeling, and Scholkopf]{schuler2013machine}
Christian~J Schuler, Harold Christopher~Burger, Stefan Harmeling, and Bernhard Scholkopf.
\newblock A machine learning approach for non-blind image deconvolution.
\newblock In \emph{CVPR}, pages 1067--1074, 2013.

\bibitem[Son et~al.(2021)Son, Lee, Cho, and Lee]{son2021single}
Hyeongseok Son, Junyong Lee, Sunghyun Cho, and Seungyong Lee.
\newblock Single image defocus deblurring using kernel-sharing parallel atrous convolutions.
\newblock In \emph{ICCV}, pages 2642--2650, 2021.

\bibitem[Su et~al.(2020)Su, Fang, Kang, Hu, Pietik{\"a}inen, and Liu]{su2020dynamic}
Zhuo Su, Linpu Fang, Wenxiong Kang, Dewen Hu, Matti Pietik{\"a}inen, and Li Liu.
\newblock Dynamic group convolution for accelerating convolutional neural networks.
\newblock In \emph{ECCV}, pages 138--155, 2020.

\bibitem[Tan and Le(2019)]{tan2019mixconv}
Mingxing Tan and Quoc~V Le.
\newblock Mixconv: Mixed depthwise convolutional kernels.
\newblock In \emph{BMVC}, 2019.

\bibitem[Timofte et~al.(2014)Timofte, De~Smet, and Van~Gool]{timofte2014a+}
Radu Timofte, Vincent De~Smet, and Luc Van~Gool.
\newblock A+: Adjusted anchored neighborhood regression for fast super-resolution.
\newblock In \emph{ACCV}, pages 111--126, 2014.

\bibitem[Vaswani et~al.(2017)Vaswani, Shazeer, Parmar, Uszkoreit, Jones, Gomez, Kaiser, and Polosukhin]{vaswani2017attention}
Ashish Vaswani, Noam Shazeer, Niki Parmar, Jakob Uszkoreit, Llion Jones, Aidan~N Gomez, Lukasz Kaiser, and Illia Polosukhin.
\newblock Attention is all you need.
\newblock \emph{NeurIPS}, 2017.

\bibitem[Wang et~al.(2019)Wang, Kan, Shan, and Chen]{wang2019fully}
Xijun Wang, Meina Kan, Shiguang Shan, and Xilin Chen.
\newblock Fully learnable group convolution for acceleration of deep neural networks.
\newblock In \emph{CVPR}, pages 9049--9058, 2019.

\bibitem[Wiener(1949)]{wiener1949extrapolation}
Norbert Wiener.
\newblock \emph{Extrapolation, interpolation, and smoothing of stationary time series: with engineering applications}.
\newblock The MIT press, 1949.

\bibitem[Wu et~al.(2022)Wu, Weng, Zhang, Wang, Yang, and Jiang]{wu2022uretinex}
Wenhui Wu, Jian Weng, Pingping Zhang, Xu Wang, Wenhan Yang, and Jianmin Jiang.
\newblock Uretinex-net: Retinex-based deep unfolding network for low-light image enhancement.
\newblock In \emph{CVPR}, pages 5901--5910, 2022.

\bibitem[Xie et~al.(2017)Xie, Girshick, Doll{\'a}r, Tu, and He]{ResNeXt}
Saining Xie, Ross Girshick, Piotr Doll{\'a}r, Zhuowen Tu, and Kaiming He.
\newblock Aggregated residual transformations for deep neural networks.
\newblock In \emph{CVPR}, pages 1492--1500, 2017.

\bibitem[Xu et~al.(2014)Xu, Ren, Liu, and Jia]{xu2014deep}
Li Xu, Jimmy~S Ren, Ce Liu, and Jiaya Jia.
\newblock Deep convolutional neural network for image deconvolution.
\newblock \emph{NeurIPS}, 27, 2014.

\bibitem[Yu and Koltun(2015)]{dilatedconvolution}
Fisher Yu and Vladlen Koltun.
\newblock Multi-scale context aggregation by dilated convolutions.
\newblock In \emph{ICML}, 2015.

\bibitem[Zhang et~al.(2017{\natexlab{a}})Zhang, Pan, Lai, Lau, and Yang]{zhang2017learning}
Jiawei Zhang, Jinshan Pan, Wei-Sheng Lai, Rynson~WH Lau, and Ming-Hsuan Yang.
\newblock Learning fully convolutional networks for iterative non-blind deconvolution.
\newblock In \emph{CVPR}, pages 3817--3825, 2017{\natexlab{a}}.

\bibitem[Zhang et~al.(2025)Zhang, Yue, Wang, Zhao, and Meng]{zhang2025blind}
Jiangtao Zhang, Zongsheng Yue, Hui Wang, Qian Zhao, and Deyu Meng.
\newblock Blind image deconvolution by generative-based kernel prior and initializer via latent encoding.
\newblock In \emph{ECCV}, pages 73--92. Springer, 2025.

\bibitem[Zhang et~al.(2017{\natexlab{b}})Zhang, Zuo, Chen, Meng, and Zhang]{dncnn}
Kai Zhang, Wangmeng Zuo, Yunjin Chen, Deyu Meng, and Lei Zhang.
\newblock Beyond a gaussian denoiser: Residual learning of deep cnn for image denoising.
\newblock \emph{IEEE TIP}, 26\penalty0 (7):\penalty0 3142--3155, 2017{\natexlab{b}}.

\bibitem[Zhang et~al.(2020{\natexlab{a}})Zhang, Gool, and Timofte]{zhang2020deep}
Kai Zhang, Luc~Van Gool, and Radu Timofte.
\newblock Deep unfolding network for image super-resolution.
\newblock In \emph{CVPR}, pages 3217--3226, 2020{\natexlab{a}}.

\bibitem[Zhang et~al.(2021)Zhang, Li, Zuo, Zhang, Van~Gool, and Timofte]{zhang2021plug}
Kai Zhang, Yawei Li, Wangmeng Zuo, Lei Zhang, Luc Van~Gool, and Radu Timofte.
\newblock Plug-and-play image restoration with deep denoiser prior.
\newblock \emph{IEEE TPAMI}, 44\penalty0 (10):\penalty0 6360--6376, 2021.

\bibitem[Zhang et~al.(2020{\natexlab{b}})Zhang, Lo, and Lu]{zhang2020high}
Pengfei Zhang, Eric Lo, and Baotong Lu.
\newblock High performance depthwise and pointwise convolutions on mobile devices.
\newblock In \emph{AAAI}, pages 6795--6802, 2020{\natexlab{b}}.

\bibitem[Zhao et~al.(2016)Zhao, Wei, Basarab, Dobigeon, Kouam{\'e}, and Tourneret]{zhao2016fast}
Ningning Zhao, Qi Wei, Adrian Basarab, Nicolas Dobigeon, Denis Kouam{\'e}, and Jean-Yves Tourneret.
\newblock Fast single image super-resolution using a new analytical solution for $\ell2$-$\ell2$ problems.
\newblock \emph{IEEE TIP}, 25\penalty0 (8):\penalty0 3683--3697, 2016.

\end{thebibliography}
